\documentclass[final,1p,times]{elsarticle}

\usepackage{lineno,hyperref}
\usepackage{amssymb,amsmath, amsthm}
\usepackage{svg}
\usepackage{multirow}
\usepackage{tabularx}
\usepackage{algorithm}
\usepackage{algpseudocode}
\usepackage{subcaption}

\journal{Null}

\begin{document}
	
	\begin{frontmatter}
		
		
		
		\title{Robust Regression with Highly Corrupted Data via Physics Informed Neural Networks}
		
		
		\author[inst1]{Wei Peng}
		\ead{weipeng0098@126.com}
		\affiliation[inst1]{organization={Defense Innovation Institute, Chinese Academy of Military Science},
			city={Beijing},
			postcode={100071}, 
			country={China}}

		\author[inst1]{Wen Yao\corref{cor}}
		\ead{wendy0782@126.com}
		\cortext[cor]{Corresponding Author}
		\author[inst1]{Weien Zhou}\ead{weienzhou@nudt.edu.cn}
		\author[inst1]{Xiaoya Zhang}
		
		\author[inst2]{Weijie Yao}
		
		\affiliation[inst2]{organization={College of Aerospace Science and Engineering, National University of Defense Technology},
			city={Changsha},
			country={China}}
		
		\begin{abstract}

        Physics-informed neural networks (PINNs) have been proposed to solve two main classes of problems: data-driven solutions and data-driven discovery of partial differential equations. 
        This task becomes prohibitive when such data is highly corrupted due to the possible sensor mechanism failing. We propose the Least Absolute Deviation based PINN (LAD-PINN) to reconstruct the solution and recover unknown parameters in PDEs – even if spurious data or outliers corrupt a large percentage of the observations. 
        To further improve the accuracy of recovering hidden physics, the two-stage Median Absolute Deviation based PINN (MAD-PINN) is proposed, where LAD-PINN is employed as an outlier detector followed by MAD screening out the highly corrupted data. Then the vanilla PINN or its variants can be subsequently applied to exploit the remaining normal data.
        Through several examples, including Poisson's equation, wave equation, and steady or unsteady Navier-Stokes equations, we illustrate the generalizability, accuracy and efficiency of the proposed algorithms for recovering governing equations from noisy and highly corrupted measurement data.

		\end{abstract}
		
		
		
		
	\end{frontmatter}
	
	
	\section{Introduction}
 
	Recently, with the explosive growth of available data and computational resources, deep learning has attracted tremendous attentions in recent years in the field of computational mechanics due to its powerful capability in nonlinear modeling of complex spatiotemporal systems. 
	One seminal work in this direction is the Physics-Informed Neural Network (PINN)\cite{raissiPhysicsinformedNeuralNetworks2019} approach introduced in a series of papers, which parameterizes a PDE’s solution as a neural network. 

	The general framework of PINN demonstrated its capacity in modeling complex physical systems such as solving/identifying PDEs.
    PINN can also exploit the available measurement data, making it possible to discover the systems whose physics are not fully understood. 
	In particular, contributions of using PINN to model various physics have been made recently. 
	For example, the Navier-Stokes equations are incorporated with data to reveal hidden fluid dynamics \cite{raissiHiddenFluidMechanics2020, caiPhysicsinformedNeuralNetworks2021a} via PINN. 
    Jin et al. \cite{jinNSFnetsNavierStokesFlow2021} employed PINN to model incompressible flow from laminar to turbulent. 
    Sun et al. \cite{sunSurrogateModelingFluid2020a} proposed PINN-based surrogates of fluids in an unsupervised manner. 
    Haghighat et al. \cite{haghighatPhysicsinformedDeepLearning2021} used PINN for inversion and surrogate modeling in the field of solid mechanics. 
    Cai et al. \cite{caiPhysicsInformedNeuralNetworks2021} investigated solutions to various prototype heat transfer problems. 
    He et al. \cite{hePhysicsinformedDeepLearning2021} studied the problem of direct and inverse heat conduction in materials.
    Lu et al. \cite{luPhysicsInformedNeuralNetworks2021} used PINN in reverse design to solve the holographic problem in optics and the fluid problem of Stokes flow.




	With $L_2$ loss being one of the most popularly used metric, the vanilla PINN calculates the squared $L_2$ norm of the PDE and initial/boundary condition residuals on the domains, respectively.	By defining differentiable loss functionals that measure how well the model fits PDEs and boundary conditions, the PINN parameters can be optimized using gradient-based approaches.
    A series of extensions to the vanilla PINN has been proposed to improve the accuracy and efficiency of handling increasingly challenging problems. Such extensions include, but are not limited to, tuning of loss weights \cite{wangUnderstandingMitigatingGradient2021, mcclennySelfAdaptivePhysicsInformedNeural2020,xiangSelfadaptiveLossBalanced2022, liuDualDimerMethodTraining2021}, novel network architectures and activation functions \cite{ramabathiranSPINNSparsePhysicsbased2021a, wangRespectingCausalityAll2022, yuGradientenhancedPhysicsinformedNeural2022, raynaudModalPINNExtensionPhysicsinformed2022}, domain decomposition methods \cite{mengPPINNPararealPhysicsinformed2020, karniadakisExtendedPhysicsInformedNeural2020}, and collocation generation algorithms \cite{dasStateoftheArtReviewDesign2022, nabianEfficientTrainingPhysics2021a, luDeepXDEDeepLearning2021, zhaoSolvingAllenCahnCahnHilliard2021,pengRANGResidualbasedAdaptive2022}.

	In the view of regression analysis, vanilla PINN is a nonlinear extension to the Ordinary Least Square (OLS) method, which we also refer to as OLS-PINN. 
	Although previous PINN methods have demonstrated promising results by combining nonlinear regression with physics-informed regularizers,  it is still absent for studies and improvements of PINN on highly corrupted data.	
    Unfortunately, like OLS in linear analysis, the data-driven part of vanilla PINN (OLS-PINN) also suffers from highly corrupted data and thus fails to reconstruct solutions or parameters correctly.
    Fortunately, in the field of robust linear analysis, the M-estimator method has been used to relieve the data corruption issue, inspired by which, we attempt to develop robust PINNs for highly corrupted observation data.

    This paper focuses on applying PINN to inverse problems, especially in the cases of incomplete boundary conditions, where solutions or parameters of the PDE equation are required to be recovered from observations that may be highly corrupted. The contributions of this paper are as follows.
    \begin{enumerate}
	\item 
   We propose the LAD-PINN model inspired by the traditional LAD (Least Absolute Derivation) approach,
   where the $L_1$ loss replaces the squared $L_2$ data loss in OLS-PINN. A systematic numerical comparison is performed on various cases of multiple data corruption types, including anomalous and spurious data. It is found that the LAD-PINN is much more robust in recovering solutions from highly corrupted observations compared to the vanilla counterpart.	
   \item Motivated by the multi-stage robust linear regression methods such as REWLS or RANSAC\cite{fischlerRandomSampleConsensus1981}, we further propose a two-stage algorithmic framework using LAD-PINN as the outlier detector. In particular, we propose a data cleaning process based on mean absolute derivation (MAD) and then solve PDEs or reconstruct solutions on the cleaned data.
    \item PINN is a simple and practical approach to fuse data with physics, the popularity of which may come from its simplicity.
    Therefore, in this paper, the introduction of additional hyperparameters is minimized in the algorithm design, and even for the two-stage algorithm, only one interpretable hyperparameter is introduced. 
    The number of remaining parameters is kept consistent with vanilla PINN. 
    The work in this paper can be easily reproduced and applied to various fields, and the code for reproducing the experiments is available on \url{https://www.github.com/weipengOO98/robust-PINN}.
    \end{enumerate}
	\textbf{Related work}.
	One class of studies for solving or reconstructing equations dealing with noise is PINNs based on Bayesian methods.
	However, these studies require probabilistic modeling of noises, i.e., it requires assumptions about the likelihood. The current works usually assume Gaussian noise and do not consider anomalous data of unknown distribution. Some enlightening works are closely related to this paper among the deterministic approaches.
 	Wang et al.\cite{Wang2022IsLP} doubted the plausibility of $L_2$ loss and noticed the impact of loss functions on training PINN. Therefore, they used $L_\infty$-norm to penalize the collocation points with more significant residuals in the PDE term or boundary condition term and used these points as adversarial samples for accelerating training. In contrast, our paper assumes that there may be unreliable observations and adopts the $L_1$-norm to reduce the impact of these outliers. Of course, these two methods are not mutually exclusive, where the $L_1$-norm can be used in the data loss term while the $L_\infty$-norm can be used in the PDE loss term.
    \cite{tranExactRecoveryChaotic2017} reconstructs the dynamics from highly corrupted observations in the dynamical system, and they also used the $L_1$-norm to deal with unreliable observations. However, their model is essentially a sparse generalized linear model, while the PINN used in this paper is highly nonlinear, resulting in substantial differences in modeling and solving.
    Peng et al.\cite{pengIDRLnetPhysicsInformedNeural2021a} mentioned examples of robust coefficient recovery by employing the $L_1$ norm, but they lacked a more comprehensive evaluation of the method. The present paper provides a systematic comparison of multiple scenarios for a variety of anomalous data as well as a two-stage framework for more accurate reconstruction. 

    The paper is organized as follows. In Section 2, the backgrounds of PINN and robust linear regression are first presented. Section 3 introduces the LAD-based robust PINN method and a two-stage framework. Section 4 compares the methods systematically through four experiments, including two examples with known PDE parameters for solution reconstruction and two examples with unknown PDE parameters for recovery. Section 5 summarizes the results and concludes the paper.
	\section{Preliminaries}
    This paper is mainly based on two fundamental techniques, one of which is PINN and the other is M-estimator derived from robust statistical analysis. These two techniques are briefly described in this section.
    
	\subsection{Physics informed neural networks}
	In this subsection, we present an brief overview of vanilla PINN (OLS-PINN) for recovering solutions and identifying unknown parameters of PDEs. 
    The main goal is to approximate the solution $u$ for the following differential equation:
	\begin{align*}
		\mathcal{N}(t,x,u;\lambda)=0, &\quad t\in[0,T],x\in \Omega,
	\end{align*}
	where $\mathcal{N}$ denotes a general differential operator that may consist of derivatives, linear terms, nonlinear terms, and possibly unknown parameters $\lambda$. $x$ denotes a spatial vector, and $t$ is a scalar about time. 

    Following the work\cite{raissiPhysicsinformedNeuralNetworks2019} of Raissi et al., given some observed spatiotemporal data $\mathbf{D}_{u}=\{(t_i, x_i, u_i)\}$, we are concerned with two problems, one of which is to recover the solution $u$, and the other is to identify $\lambda$ for the case of unknown parameters included. 
    PINN employs a feedforward neural network $\hat u$ as the surrogate for the solution. We consider $\hat u_\theta$ a classic fully-connected neural network parameterized by $\theta$. Define the residual networks, which share the same network parameters $\theta$:
	\begin{align*}
		r_{pde}(t,x;\theta,\lambda)&:=\mathcal{N}(t, x,\hat u(t,x;\theta);\lambda),
	\end{align*}
	where automatic differentiation techniques compute all derivatives.

	The shared network parameters $\theta$ are trained by minimizing loss terms that penalize the residuals for not equaling zero:
	\begin{align}\label{eq:vanilla_pinn}
		L(\theta,\lambda)&=\omega\cdot L_{pde}(\theta,\lambda)+L_u(\theta),\quad(\omega>0)
	\end{align}
	where $L_{pde}$ is the PDE loss term, $L_u$ is the data loss term, and $\omega>0$ is a user-defined weighting coefficient for PDE loss. The PDE loss term is approximated by sampling inside the domain $\Omega$:
	\begin{align}\label{eq:pde_l}
	    L_{pde}(\theta,\lambda)&=\frac{1}{\#\mathbf{D}_{pde}}\sum_{(x,t)\in \mathbf{D}_{pde}} |r_{pde}(t,x;\theta,\lambda)|^2,
	\end{align}
	where $\mathbf{D}_{pde}$ is the collocation set for enforcing strong forms of PDEs, and the operator $\#$ denotes the number of elements in a set. 
    For OLS-PINN, the data loss term is constructed by measuring the difference between the predicted values and given labels with the squared $l_2$-norm:
	\begin{align*}
	    L_{u}(\theta)&=\frac{1}{\#\mathbf{D}_{u}}\sum_{(x_i,t_i,u_i)\in \mathbf{D}_u} |\hat u(t_i, x_i;\theta)-u_i|^2,
	\end{align*}
	 where $\mathbf{D}_u=\{(t_i, x_i, u_i)\}$ may contain samples from initial conditions, boundary conditions, and noisy measurements. 
	 Finally, the solution $\hat u_\theta$ and the unknown PDE parameter $\lambda$ are approximated by minimizing the total loss $L(\theta,\lambda)$:
	\begin{align}\label{eq:minimize}
		\theta^\ast,\lambda^\ast=\arg\min_{\theta,\lambda} L(\theta, \lambda).
	\end{align}
    The minimization problem can be solved by the gradient descent method Adam \cite{kingmaAdamMethodStochastic2015} and a limited memory variant of BFGS optimizer(L-BFGS-B) \cite{byrdLimitedMemoryAlgorithm1995}.
	
	\subsection{L1 robust regression for linear models}

    We consider recovering PDE solutions and unknown parameters from highly corrupted data that may contain outliers. 
    Since methods of handling atypical data are comprehensively studied in robust linear regression analysis, which deals with outliers in the dataset simultaneously with the fitting process. 
    Robust linear regression analysis provides motivations and approaches, allowing us to speculate on similar problems in data-driven PINN-related tasks. 
    We will use some of these concepts from robust linear analysis contexts.
    Therefore, this subsection will briefly introduce several robust M-estimators for robust linear regression that are closely related to this paper, and the proposed robust PINN is essentially an extension to the nonlinear fields in the sense of deep neural network models. 

	Given the observations $(x_i, u_i)$, $i = 1,\cdots,n$, in order to understand how the responses $u_i$s are related to the covariates $x_i$s, linear models usually assume the following form:
	\begin{align*}
	    u_i=x_i^T\theta +\varepsilon,
	\end{align*}
	where $\theta\in\mathbb{R}^p$ is unknown and to be determined. For the Gaussian noise $\varepsilon$, choosing $\rho_i(\cdot):=\|\cdot\|^2$ is preferred since the estimator coincides with maximum-likelihood estimates with the assumption of independent normally distributed errors with $0$ mean and equal variances. 
    The deduced least squared form can be solved readily with well-understood linear numerical algebra methods \cite{demmelAppliedNumericalLinear1997}:
	\begin{align}\label{eq:ols}
	    \theta^\ast := \arg\min_\theta \sum_{i=1}^n(u_i-x_i^T\theta)^2.
	\end{align}
    The Ordinary Least Square (OLS) estimator introduced above is not robust since the occurrence of even a single outlier can spoil the result completely \cite{rousseeuwRobustRegressionMeans1984}.

\begin{figure}[htbp]
  \centering
  \includegraphics[width=0.9\textwidth]{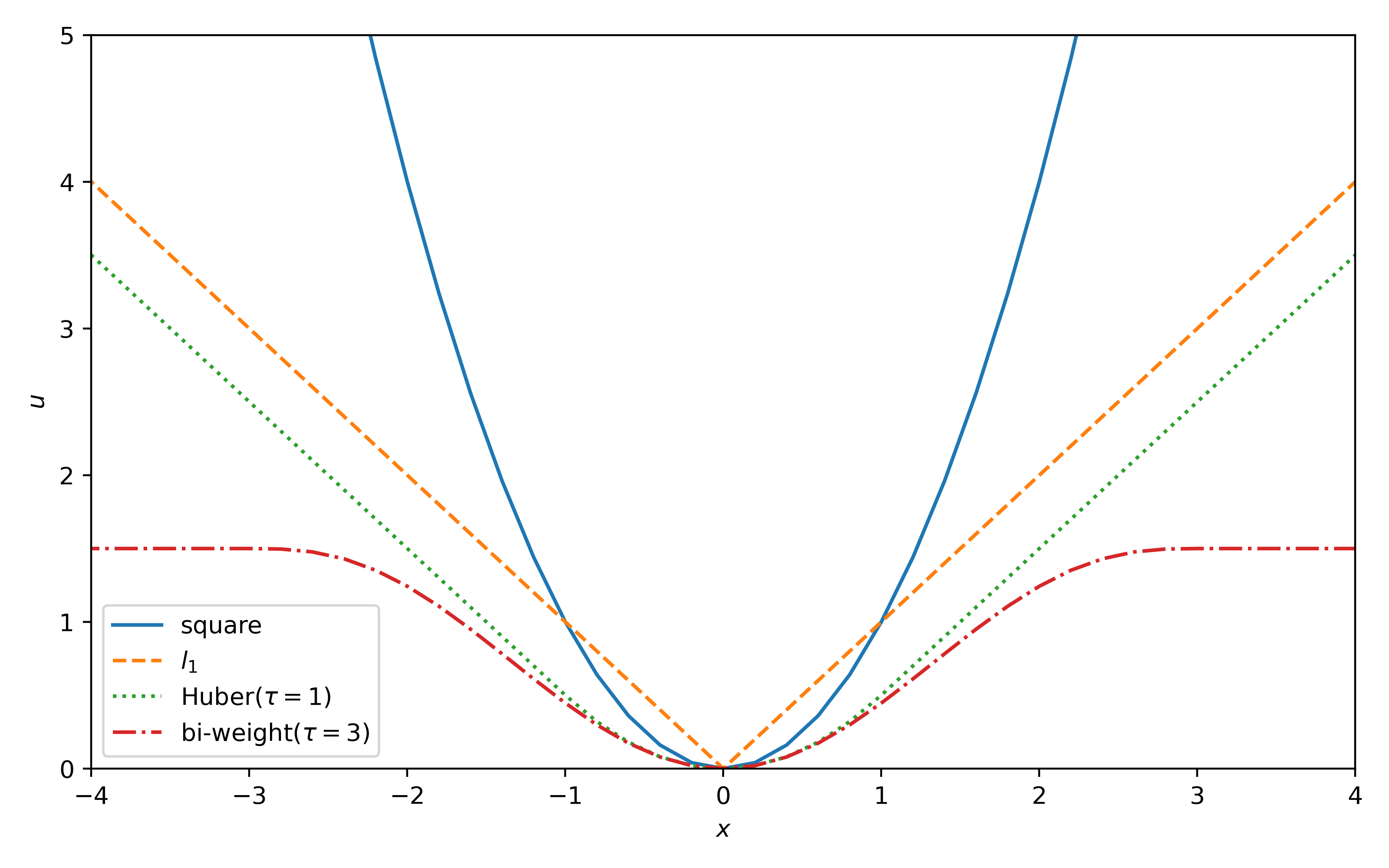}
  \caption{Different loss functions. Curve shapes of Huber's loss function and bi-weight are controlled by a hyperparameter $\tau$. The square loss increases the fastest when $|x|$ becomes large.}
  \label{fig:lfuncs}
\end{figure}

	By replacing the least squares criterion \eqref{eq:ols} with a more robust criterion, the M-estimate \cite{huberRobustEstimationLocation1964} of $\theta$ is
	\begin{align*}
	    \theta^\ast:=\arg\min_\theta\sum\rho_i\left(\frac{u_i- x_i^T\theta}{\hat \sigma}\right),
	\end{align*}
    where the choice of $\rho_i$ has a significant effect on the results, and $\hat \sigma$ is the scale estimate of the error. When $\rho_i=\|\cdot\|^2$, it becomes OLS. Consider other functions such as $L_1$ norm, Tukey's biweight function, or Huber's loss function. The schematic of the $4$ loss functions is shown in Fig.\ref{fig:lfuncs}. The three methods other than OLS can usually handle outliers. The reason is that these methods give much less weight to large deviations. For more tens of robust estimator loss functions, please refer to \cite{demenezesReviewRobustMestimators2021}. 

In particular, the $L_1$ loss minimizes the sum of the absolute errors and is therefore also known as the Least Absolute Derivation (LAD) method:
\begin{align}
    \min_\theta \sum\left|u_i-x_i^T\theta\right|.
\end{align}

Although the LAD is inefficient in the robust linear regression context, i.e., LAD estimates have a low efficiency of 0.64 when the errors are distributed normally\cite{huberRobustEstimationLocation1964}, LAD has the most straightforward form among all M-estimators and is straightforward for generalization. 
In the next section, we will use LAD to constrain the data items in PINN and illustrate the potential advantages and shortcomings of LAD compared to other M-estimators in PINN applications.

\section{LAD Physics-Informed neural networks}

Not only is OLS susceptible to outliers, but nonlinear models/neural networks may also be sensitive to outliers in datasets. 
For example, data poisoning attacks can cause specific neural network errors by adding spurious data to the training dataset \cite{steinhardtCertifiedDefensesData2017}. 
In regards to the implementation of vanilla PINN, even one atypical observation can cause significant errors in the results (Fig.\ref{fig:poisson_breaker}). 
Therefore, motivated by the LAD method from robust linear analysis, we ameliorate this problem by minimizing the absolute value of the data discrepancy to handle possibly highly corrupted data. 
Then the PDE constrained optimization model is formulated as:
\begin{align*}
    \min_{\theta, \lambda} &\quad \sum_{(t_i,x_i,u_i)\in\mathbf{D}_u} |u_i-u_\theta(t_i, x_i;\lambda)|\\
    s.t. &\quad \mathcal{N}(t,x,u_\theta,\lambda)=0,
\end{align*}
which is then transformed into the following unconstrained form:
\begin{align}\label{eq:uncon_pde}
   \textbf{(LAD-PINN)}:\quad \min_{\theta, \lambda} &\quad \frac{1}{\#\mathbf{D}_u}\sum_{(t_i,x_i,u_i)\in\mathbf{D}_u} |u_i-u_\theta(t_i, x_i;\lambda)|+\omega\cdot L_{pde}.
\end{align}

When the measurement $u_i$ is a vector instead of a scalar, the data loss $\sum_i |e_i|$ extends to $\sum _i \|e_i\|_1$ or $\sum _i \|e_i\|_2$, which corresponds to the sparse regularization and the group sparse regularization \cite{yuanModelSelectionEstimation2006}, respectively. In our experiment involving vector measurements, we choose $\sum _i \|e_i\|_1$ as the data loss. 

Eq.\eqref{eq:uncon_pde} replaces the squared term in Eq.\eqref{eq:vanilla_pinn} with the $L_1$ loss term. Note that the unconstrained model is not equivalent to the PDE constrained model. Indeed, if a physical law is only constrained approximately, the solution may exhibit a different quality or even reach a completely wrong answer. However, \textit{All models are wrong, but some are useful} \cite{boxNonNormalityTestsVariances1953}. 
Since unconstrained optimization models are usually easier to be solved, Eq.\eqref{eq:vanilla_pinn} has been widely used in various fields along with the vanilla PINN. Therefore, the unconstrained LAD-PINN model (Eq.\eqref{eq:uncon_pde}) also promises to achieve satisfactory results.

From the perspective of solving a constrained optimization, using the penalty function method or the augmented Lagrange multiplier method seems to be more appropriate \cite{demenezesReviewRobustMestimators2021,luPhysicsInformedNeuralNetworks2021,basirPhysicsEqualityConstrained2022}. However, these two kinds of methods not only involve more hyperparameter selections but also require solving the minimization sub-problems several times. In order to reduce burdens in the endless algorithm hyperparameter selection, we insist on converting the PDE constraint into an unconstrained minimization form. The proposed method only needs to select one hyperparameter at the modeling level and solve the minimization problem only once.

In Eq.\eqref{eq:uncon_pde}, we are trying to minimize the absolute derivations instead of squared errors. Similar to LAD used in robust linear regression, one can deduce that the $L_1$ norm loss function is less likely to be affected by outliers compared to the squared $l_2$ norm that is employed in vanilla PINN. 
Hence, the impact of a significant deviation of observed data from predicted values has been greatly diminished.

Among the various methods of robust linear analysis, although the M-estimator LAD has only a breakdown point and a low efficiency of 0.64 when the errors are distributed normally \cite{rousseeuwRobustRegressionMeans1984}, it is free from additional hyperparameter selection. 
Since training a PINN is usually much more expensive than solving a linear model. Currently, we may not have an efficient way to select hyperparameters and evaluate them. Therefore, in this paper, LAD is chosen to be combined with PINN. 
In addition, it is also popular to use the $L_1$ loss term for weight regularization in deep neural networks, and thus the corresponding available optimizer for the non-smooth problem seems more applicable to LAD-PINN than the models deduced from other delicate robust estimators.

\subsection{LAD-PINN as an abnormal detector: a two-stage framework}

Data cleaning is a mandatory step in the usual data-driven tasks. Although LAD-PINN completes both anomalous data processing and subsequent training tasks simultaneously, we can use LAD-PINN only for data cleaning to get a relatively clean data set.
LAD-PINN can usually recover stable results from highly corrupted data, but due to the lower efficiency of LAD relative to OLS, the accuracy of LAD-PINN is usually lower than vanilla PINN in the noiseless or low standard Gaussian noise cases, which may result in the failures in revealing hidden physics (subsection \ref{sec:steady_pressure} and subsection \ref{sec:unsteady_pressure}).

Referring to the idea of the iterative robust regression analysis methods such as REWLSE\cite{gerviniClassRobustFully2002} or RANSAC\cite{fischlerRandomSampleConsensus1981} which gradually eliminates anomalies through a multi-stage approach, considering the iterative multi-stage might bring heavy computational burdens for training PINN, we propose a two-stage training method in moderation. 

In the first stage, the target physical field is stably reconstructed by LAD-PINN under potentially highly corrupted data. LAD-PINN works as an anomaly detector, and its predicted field is regarded as a reference to identify outliers based on the distribution of the difference between observed and predicted values in the dataset. The identified data points are removed to obtain a cleaned dataset. The physical field is then reconstructed in the second stage based on the cleaned dataset using OLS-PINN. The network trained by LAD-PINN in the first stage is also used as a pre-trained network to help accelerate the training of the second stage. 
This two-stage framework is described by Alg.\ref{alg:two-PINN}.

\begin{algorithm}
	\caption{A two-stage robust PINN framework}\label{alg:two-PINN}
	\begin{algorithmic}[1]
		\Require Dataset with possibly highly corrupted observations: $\mathbf{D}_u=\{(t_i,x_i,u_i)\}$
		\Statex PDE: $\mathcal{N}(t,x,u;\lambda)=0$
        \Statex Filter parameter: $k$
		\State The first stage training by LAD-PINN;
		\Statex \begin{align}\hat u^+,\lambda^+ \gets \arg\min_{\hat u, \lambda} \frac{1}{\#\mathbf{D}_u}\sum_{(t_i,x_i,u_i)\in\mathbf{D}_u}|\hat u(t_i,x_i)-u_i|+\omega\cdot \|\mathcal{N}(t,x,\hat u;\lambda)\|_2^2
			\end{align}
		\State Screening out the highly corrupted data with large $|\hat u^+(t_i,x_i)-u_i|$;
		\Statex 
		\begin{align}\mathbf{\hat D_u}\gets Filter(\mathbf{D}_u,\hat u^+;k)\end{align}
		\State The second stage training by OLS-PINN;
		\Statex 
		\begin{align}\hat u^\ast, \lambda^* \gets \arg\min_{\hat u, \lambda} \frac{1}{\#\mathbf{\hat D}_u}\sum_{(t_i,x_i,u_i)\in\mathbf{\hat D}_u}|\hat u(t_i,x_i)-u_i|+\omega \cdot\|\mathcal{N}(t,x,\hat u;\lambda)\|_2^2
		\end{align}
		\Statex\Return $\hat u^\ast$, $\lambda^\ast$
	\end{algorithmic}
\end{algorithm}



The two minimization problems in Line 1 and Line 3 in Alg.\ref{alg:two-PINN} are solved following the routines of PINN. Various acceleration techniques can also be integrated into the framework, including but not limited to the improvements in the introduction.

Line 2 detects and filters the abnormal data. Suppose the original dataset is $\mathbf{D}_u=\{(t_i, x_i, u_i)\}$ and the set of corresponding predicted values obtained using the LAD-PINN model is $\mathbf{D}_{p}=\{t_i, x_i, \hat u^+(t_i, x_i)\}$. 
Outliers are determined by comparing the differences between the two data sets, i.e., $\{t_i, x_i, \hat u^+(t_i, x_i)-u_i\}$. Accuracy of this approach may heavily depend on knowledge of the prior error distribution, and accurate priors may significantly enhance the results. Here we provide two approaches in the absence of specific priors.

In the first approach, the points corresponding to a certain percentage $k$ of the maximum absolute errors are designated anomalies and removed. This approach is simple and may be very effective especially if there is an accurate prior for the proportion. 
Otherwise, when the proportion $k$ is chosen too large, the normal data with an acceptable scale of noises may be removed mistakenly, and these removed normal data may be concentrated on a poorly trained region, resulting in an observation hole, as Fig.\ref{fig:piv_ratio_remove}(d) shows. 
When the proportion $k$ is chosen too small, the small amount of remaining abnormal data can also cause the subsequent OLS-PINN to fail due to its high sensitivity to outliers.

Another approach is to assume that the differences $\{u_i-\hat u^+(t_i,x_i)\}$ of non-outliers obey a normal distribution. We estimate the standard deviation of the distribution by median absolute deviation (MAD) and use the standard deviation to construct a threshold to exclude the outliers. The idea of MAD was (re-)discovered and popularized by \cite{jennisonRobustStatisticsApproach1987}. Since the MAD estimation is very robust with a breakpoint of 50\%\cite{leysDetectingOutliersNot2013}, the variance of the normal distribution can be estimated in a highly corrupted dataset to screen out the outliers. Specifically, the estimated standard deviation is
\begin{align*}
\hat \sigma_e = \frac{1}{1.6777}\cdot\textrm{median}(|u_i-\hat u^+(t_i,x_i)|).
\end{align*}
If $u_i$ lies in the neighborhood $\{u|~|\hat u^+(t_i,x_i)-u|\leq k\cdot\hat\sigma_e\}$, the point is reserved; otherwise, it is identified as abnormal and removed. $k$ is usually chosen as $2.0$, $2.5$ or $3.0$ \cite{leysDetectingOutliersNot2013}. Subsequent experiments have shown that the method is not sensitive to $k$, and similar results can be obtained for all three common choices.

In the paper, the combination of Adam and L-BFGS-B is used for the first stage, and the second stage can be accelerated by using the pre-trained $\hat u^+$ in the first stage as an initialization. 
In the second stage of training, several steps with a large learning rate are first applied for regularization\cite{largeLR}, which can avoid falling into the local optimal obtained by the first stage, although it may temporarily reduce the accuracy of the model.

\section{Experiments}
 In the numerical experiments, we consider two types of problems separately: whether the controlling PDE contains unknown parameters. 
 
 First, we consider the problems of recovering the physical field under incomplete boundary conditions with known PDE parameters, which include solving the one-dimensional Poisson's equation and two-dimensional steady Navier-Stokes (N-S) equations. Then we consider problems containing unknown PDE parameters, including a one-dimensional wave equation and two-dimensional unsteady N-S equations. In particular, we reconstruct the hidden pressure fields using the two-stage algorithm (Alg.\ref{alg:two-PINN}) in the two N-S equation problems.

The noisy data is generated by:
\begin{align*}
u_i:=\hat u_i+\varepsilon_i
\end{align*}
where $\hat u_i$ is extracted from reference solutions, which are regarded as exact. Furthermore, we consider noise terms $\varepsilon_i$ from different random distributions.

\begin{figure}[htbp]
  \centering
  \includegraphics[width=0.9\textwidth]{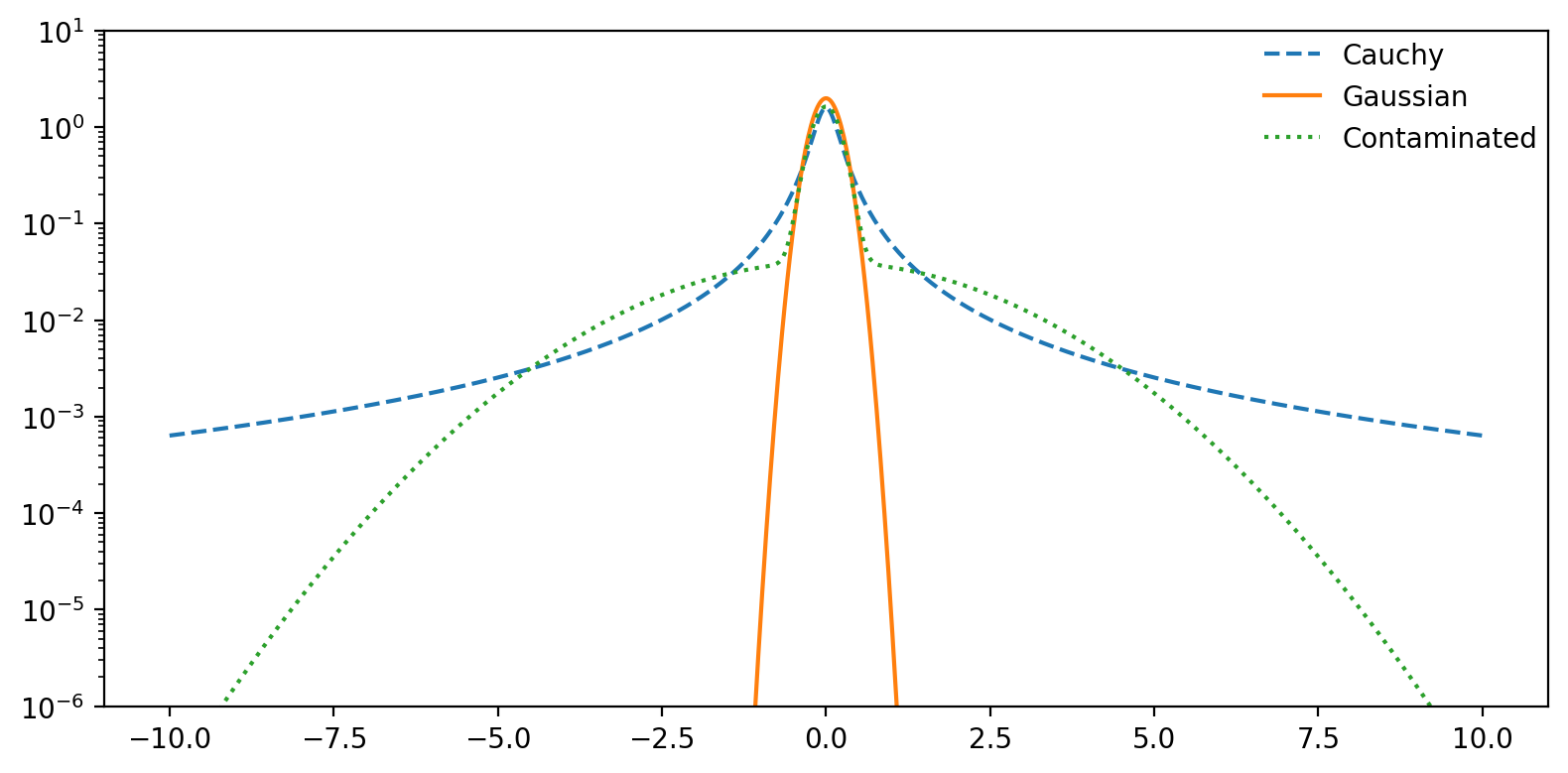}
  \caption{Probability distribution functions corresponding to different noise types. The Cauchy and mixed Gaussian distributions are more prone to produce highly corrupted data.}
  \label{fig:pfuncs}
\end{figure}

The following kinds of data corruption \cite{yuRobustLinearRegression2017} are considered in the experiment, including noises obeying Gaussian distribution, mixed Gaussian distribution, Cauchy distribution, and fixed spurious values in the $u_i$. We presume that for the reference $u$ to be corrupted, a scaling factor is first defined as $\hat\sigma=std(u)$.  To facilitate a uniform description of the different data corruption types, assume the corrupted level $\alpha\in(0,1)$ for all types. Specifically, five corruption classes are employed:
	\begin{enumerate}
        \item \textbf{Gaussian}: Noises obey normal distribution, i.e., $\varepsilon_i\sim\mathcal{N}(0,\alpha^2\hat\sigma^2)$, the standard derivation of which is $\alpha\cdot \hat\sigma$.
	    \item \textbf{Contaminated}: The noises obey Gaussian mixed distribution, generating a large number of Gaussian noise data with small variance and a small number of Gaussian noise data with significant variance, i.e., $\varepsilon\sim 0.8\cdot\mathcal{N}(0,\alpha^2\hat\sigma^2) + 0.2\cdot \mathcal{N}(0,(10\alpha\hat\sigma)^2)$. 
	    The standard derivation is $4.64\alpha\hat\sigma$. To reduce the simulation variance, we fix the ratio of noisy data from the two Gaussian distributions, and thus the noises are not independent.
	    \item \textbf{Cauchy}: The noises obey the Cauchy distribution:
	    \begin{align}
	        \varepsilon\sim \frac{1}{\pi\alpha\hat\sigma(1+\frac{x^2}{\alpha^2\hat\sigma^2})}
	    \end{align}
	    the variance of which is undefined.
	    \item \textbf{Outlier}: 
        Set $100\alpha\%$ of the observations $u_i$ to a fixed spurious value. Unlike the several noises above, this data corruption is asymmetric, 
	    \item \textbf{Mixed}: Set $100\alpha\%$ of the observations $u_i$ to a fixed spurious large value (10 or 30). The noises that obey the normal distribution $\mathcal{N}(0,\alpha^2\hat\sigma^2)$ are added to the rest of the observations.
	\end{enumerate}
 The different distribution probability density functions corresponding to the several kinds of noise are shown in Fig.\ref{fig:pfuncs}. The heavy tail usually means that noises are prone to large derivation.
 

All numerical experiments are mainly based on Tensorflow and other open-source Python packages.
 The fully connected MLP is employed in all experiments. The activation function is $tanh$, and the rest of the detailed configurations are described in the respective experiment. Unless otherwise specified, the error in the experiments is defined by the relative $l_2$ error, i.e.,
 \begin{align}
     \textrm{RE}(\hat u)=\frac{\|\hat u-u\|_2}{\|u\|_2}.
 \end{align}

Configurations for the four experiments are listed in Tab \ref{tab:configurations}.
\begin{table}
\begin{tabular}{l|c|c|c|c}
\hline
\multirow{2}{*}{PDE} & Unknown & \multirow{2}{*}{Layers}  &  Adam  &   \multirow{2}{*}{Two-stage}  \\
                     & Parameters&      &iterations&\\
\hline
\hline
Poisson &      No & [1,50,50,50,50,1] &       15000 &   No \\
Steady N-S &   No & [2,40,40,40,40,40,40,40,40,5]  &       10000 &   Yes \\
Wave &        Yes & [2,40,40,40,40,1]       &       10000 &    No \\
Unsteady N-S& Yes & [3,20,20,20,20,20,20,20,20,2]  &       10000 &   Yes \\
\hline
\end{tabular}
\caption{Configurations for each experiment: The \textit{Unknown Parameters} column indicates if there is unknown parameters in the controlling PDEs. The \textit{Layers} column describes the neuron numbers used in each layer of a neural network. The \textit{Adam iterations} column tells the maximum number of Adam iterations. The \textit{Two-stage} column shows whether the two-stage framework is employed in the experiment.}
\label{tab:configurations}
\end{table}
\subsection{One-dimensional Poisson's equation}

	The first experiment considers a simple one-dimensional Poisson's equation:
	\begin{align}\label{eq:poisson}
	    u_{xx}=-16\sin(4x).
	\end{align}
    Assume that the ground truth is $u(x)=\sin(4x)+1$.
    Observations are sampled from the interval $\left[-\pi, -\pi/2\right]$ and the solution on the interval $\left[-\pi, \pi\right]$ is to be reconstructed.

\begin{figure}[htbp]
  \centering
  \includegraphics[width=0.9\textwidth]{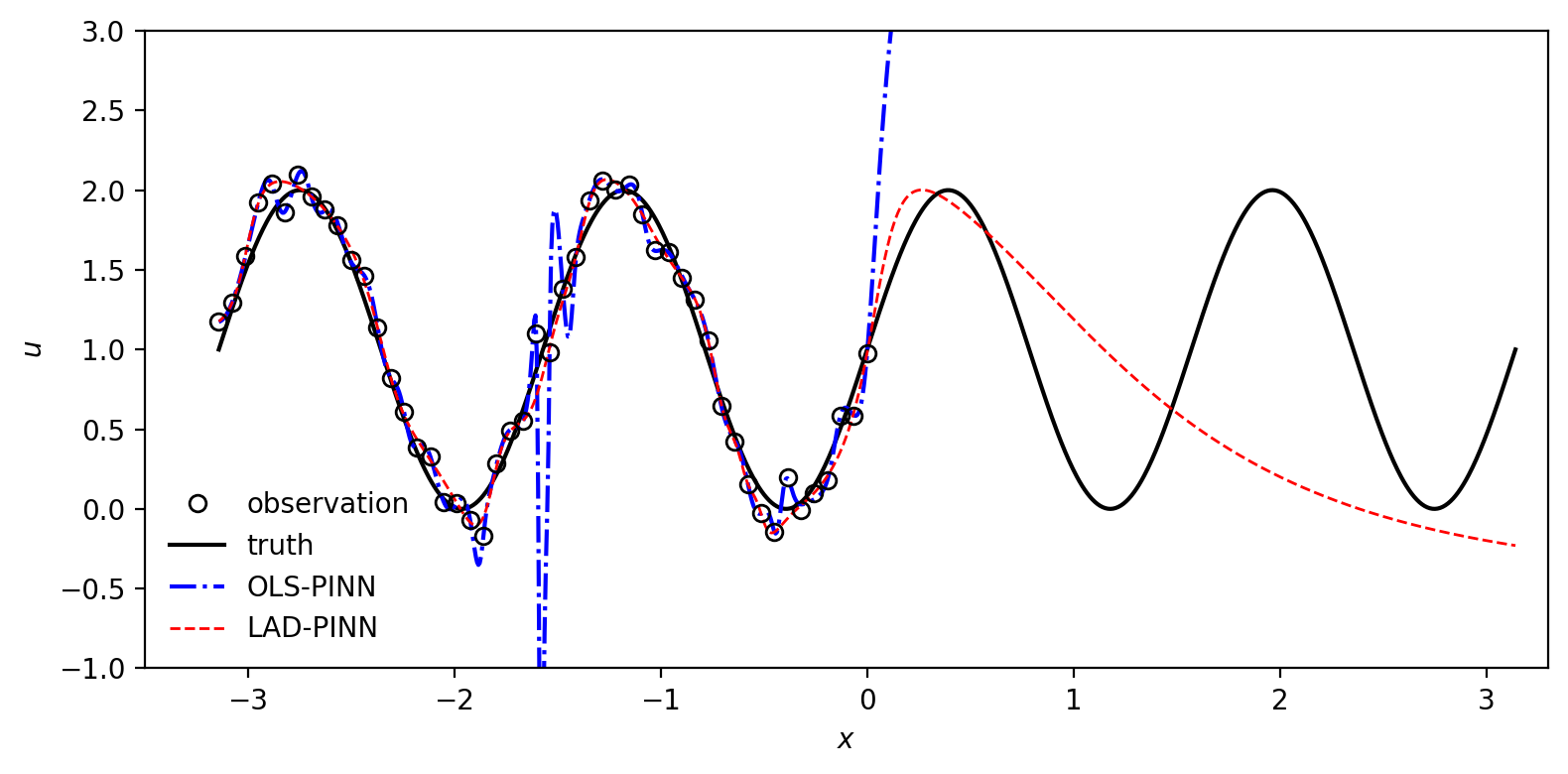}
  \caption{Regression without physics-informed regularizers. It can be challenging to generalize outside the observations, as there is no guarantee that the NN model has learned the relevant physics. OLS-NN overfits the noisy observations apparently.}
  \label{fig:poisson_no_reg}
\end{figure}
\begin{figure}[htbp]
  \centering
  \includegraphics[width=0.9\textwidth]{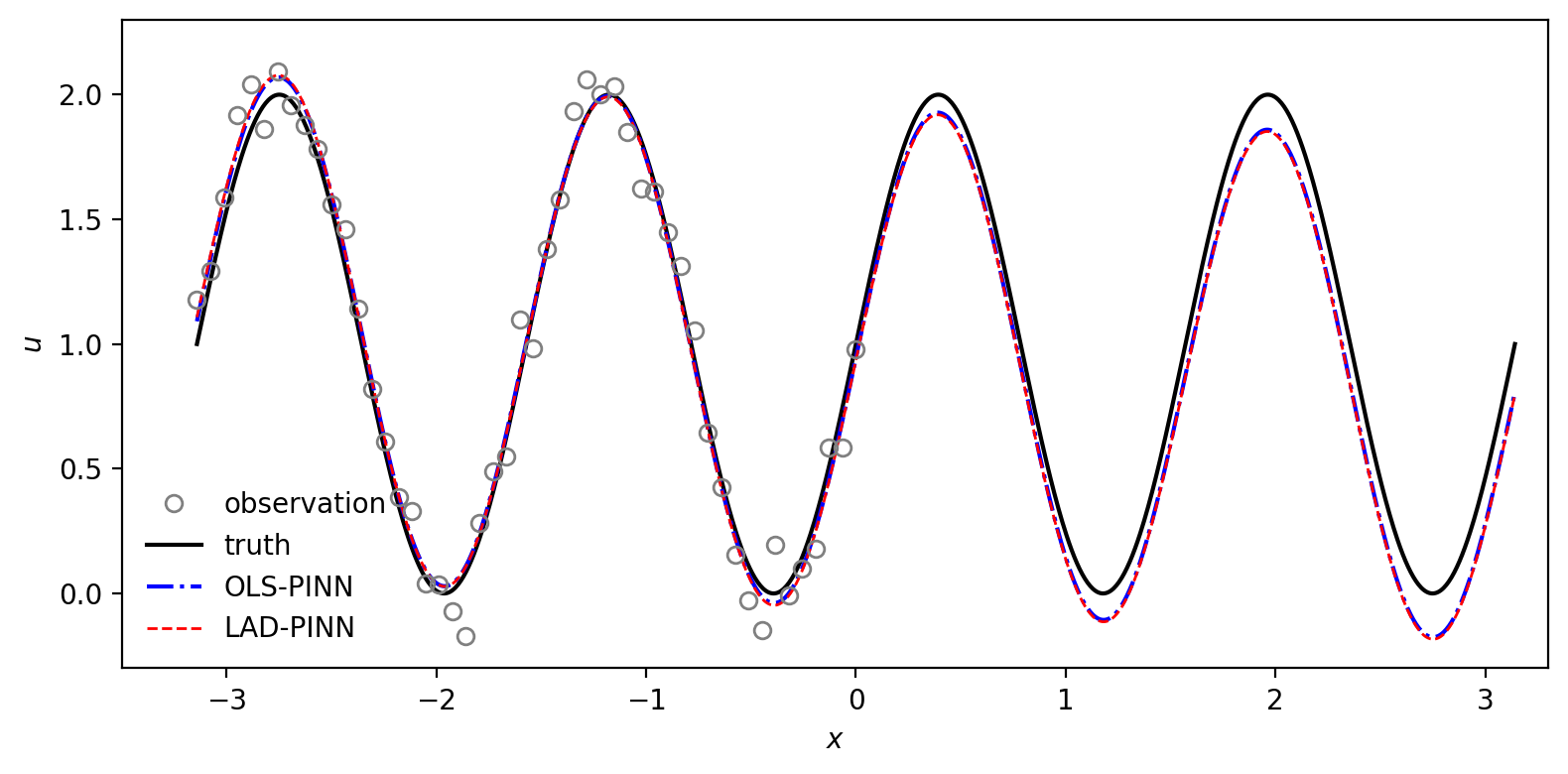}
  \caption{Nonliner regression with physics-informed regularization, i.e., PINN. When the PDE loss term is integrated, both LAD-PINN and OLS-PINN learn the solution successfully.}
  \label{fig:poisson_no_reg2}
\end{figure}

    First, we clarify that it is almost impossible to recover the solution on interval $[-\pi, \pi]$ based on observations from $[-\pi, 0]$ without physical constraints. 
    When a neural network is applied for fitting, the results is shown in Fig. \ref{fig:poisson_no_reg}. 
    In the absence of embedding physical priors, the neural network has no generalization ability in the sense of physical laws and cannot learn a correct solution on $[-\pi, \pi]$. 
    On the other hand, for noisy data, the neural network under sufficient training will overfit the data. Integrating the PDE loss can solve both problems as shown in Fig.\ref{fig:poisson_no_reg2}.

	\begin{figure}[htbp]
  \centering
  \includegraphics[width=0.9\textwidth]{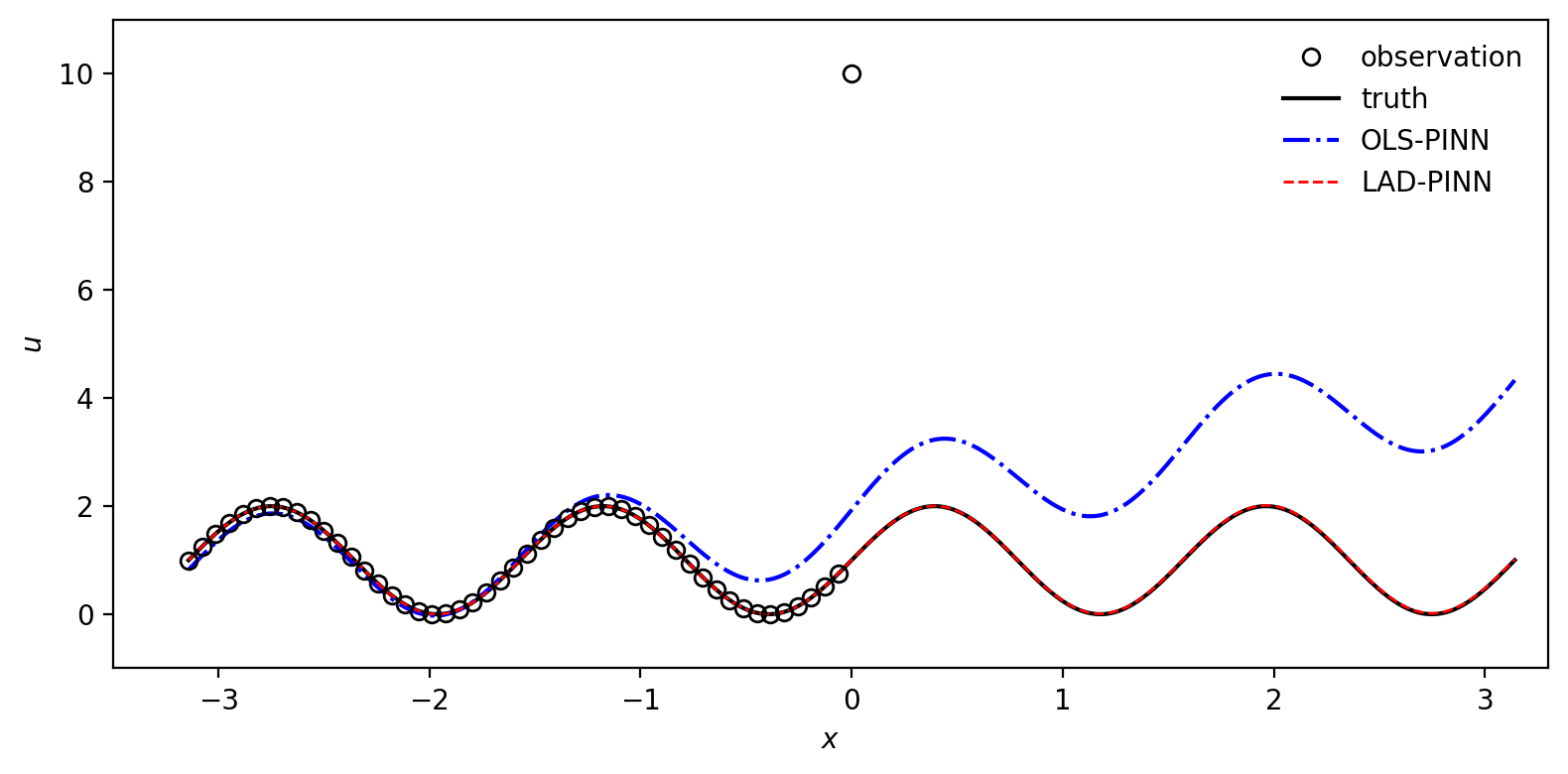}
  \caption{Observations with a single outlier in the $u$-direction. LAD-PINN succeed to recover the solution while OLS-PINN biases the result.}
  \label{fig:poisson_breaker}
\end{figure}

    Therefore, it is necessary to consider PDE loss in such regression or reconstruction tasks. Denote $\hat u$ the computed solution approximated by a neural network:
	\begin{align}
	    \hat u(x)=net(x).
	\end{align}
	Then the unconstrained optimization model is approximated on collocation points and observations:
	\begin{align}
	    L:=\omega\cdot\frac{1}{\#\mathbf{D}_{pde}}\sum_{x_i\in\mathbf{D}_{pde}} |\hat u_{xx}(x_i)|^2+\frac{1}{\#\mathbf{D}_u}\sum_{(x_i,u_i)\in \mathbf{D}_u}\left|\hat u(x_i)-u_i\right|
	\end{align}
	
 For the one-dimensional time-independent problem \eqref{eq:poisson}, the general solution is
    \begin{align}
        u(x)=\sin(4x)+ax+b.
    \end{align}
 If the observations are free of noise, only two different data points on the interval can determine the solution, and thus the solution can be recovered exactly. Therefore, only the data corruption cases are considered in the experiment.


Taking $\omega=1.0$, the number of observations is 1,000 and the fixed number of iterations is 15,000 Adam steps, followed by an L-BFGS-B optimizer. The results are shown in Tab.\ref{tab:poisson_error_u_noise}.There is an increase in the resultant error as the corruption level rises. 
Under Gaussian noise, the performance of LAD-PINN and OLS-PINN are similar, but for the case of Contaminated, Cauchy and Outliers, LAD-PINN has an advantage in accuracy, and for the cases containing spurious data in the $u$-direction, OLS-PINN can hardly reconstruct correctly.

\begin{table}
\begin{tabular}{ll|rrrrr}
\hline
$\alpha$ & Loss&  Gaussian &  Contaminated &    Cauchy &   Outlier &     Mixed \\
\hline
\hline
\multirow{2}{*}{0.10} & LAD &     0.618 &       1.702 &   2.155 &    0.029 &   2.851 \\
      & OLS &     0.658 &       4.763 &   5.706 &  108.798 & 108.691 \\
      \hline
\multirow{2}{*}{0.15} & LAD &     0.700 &       1.712 &   2.711 &    0.316 &   3.097 \\
      & OLS &     0.784 &       7.235 &   8.409 &  146.125 & 146.089 \\
      \hline
\multirow{2}{*}{0.20} & LAD &     0.821 &       1.628 &   2.758 &    0.411 &   4.018 \\
      & OLS &     1.150 &       9.586 &  11.505 &  160.362 & 159.951 \\
      \hline
\multirow{2}{*}{0.25} & LAD &     0.953 &       1.521 &   2.895 &    0.099 &   6.197 \\
      & OLS &     1.409 &      11.692 &  14.421 &  197.860 & 197.699 \\
      \hline
\multirow{2}{*}{0.30} & LAD &     1.122 &       1.539 &   3.022 &    0.187 &   9.035 \\
      & OLS &     1.698 &      14.366 &  17.163 &  213.080 & 213.321 \\
\hline
\end{tabular}
\caption{Relative errors(\%) of the solution $u$ to the Poisson's equation with different corruption levels for $\#\mathbf{D}_u=500$.}
\label{tab:poisson_error_u_noise}
\end{table}
Fixing the corruption level $\alpha$ to be $0.2$, we vary the number of given observations. As is shown in Tab.\ref{tab:poisson_error_u_size}, OLS-PINN cannot recover the solution with spurious values in the $u$-directions. Instead, LAD-PINN recovers all the solutions. Furthermore, LAD-PINN achieves better accuracy expect for the Gaussian noises, which is consistent to the high-efficiency of OLS and robustness of LAD in robust linear analysis.
\begin{table}
\begin{tabular}{ll|rrrrr}
\hline
$\#\mathbf{D}_u$ & Loss        &  Gaussian &  Contaminated &    Cauchy &   Outlier &     Mixed \\
\hline
\hline
\multirow{2}{*}{100} & LAD &    11.482 &      11.374 &   1.783 &    0.168 &   5.983 \\
    & OLS &     4.777 &      14.744 &  29.090 &  112.737 & 112.755 \\\hline
\multirow{2}{*}{200} & LAD &     1.478 &      12.855 &   3.405 &    0.292 &   3.290 \\
    & OLS &     1.772 &      28.526 &  72.338 &  221.448 & 218.376 \\\hline
\multirow{2}{*}{300} & LAD &     0.568 &       0.639 &   3.375 &    0.823 &   6.595 \\
    & OLS &     0.999 &       5.779 &  20.494 &  197.473 & 197.046 \\\hline
\multirow{2}{*}{400} & LAD &     3.293 &       2.802 &   1.252 &    1.034 &   3.259 \\
    & OLS &     3.222 &       7.912 &   5.297 &  141.417 & 138.910 \\\hline
\multirow{2}{*}{500} & LAD &     0.821 &       1.628 &   2.758 &    0.732 &   4.159 \\
    & OLS &     1.150 &       9.586 &  11.505 &  160.334 & 159.656 \\
\hline
\end{tabular}
\caption{Relative errors(\%) of the solution $u$ to the Poisson's equation with different dataset sizes for $\alpha=0.2$.}
\label{tab:poisson_error_u_size}
\end{table}

\begin{figure}[htbp]
  \centering
  \includegraphics[width=0.9\textwidth]{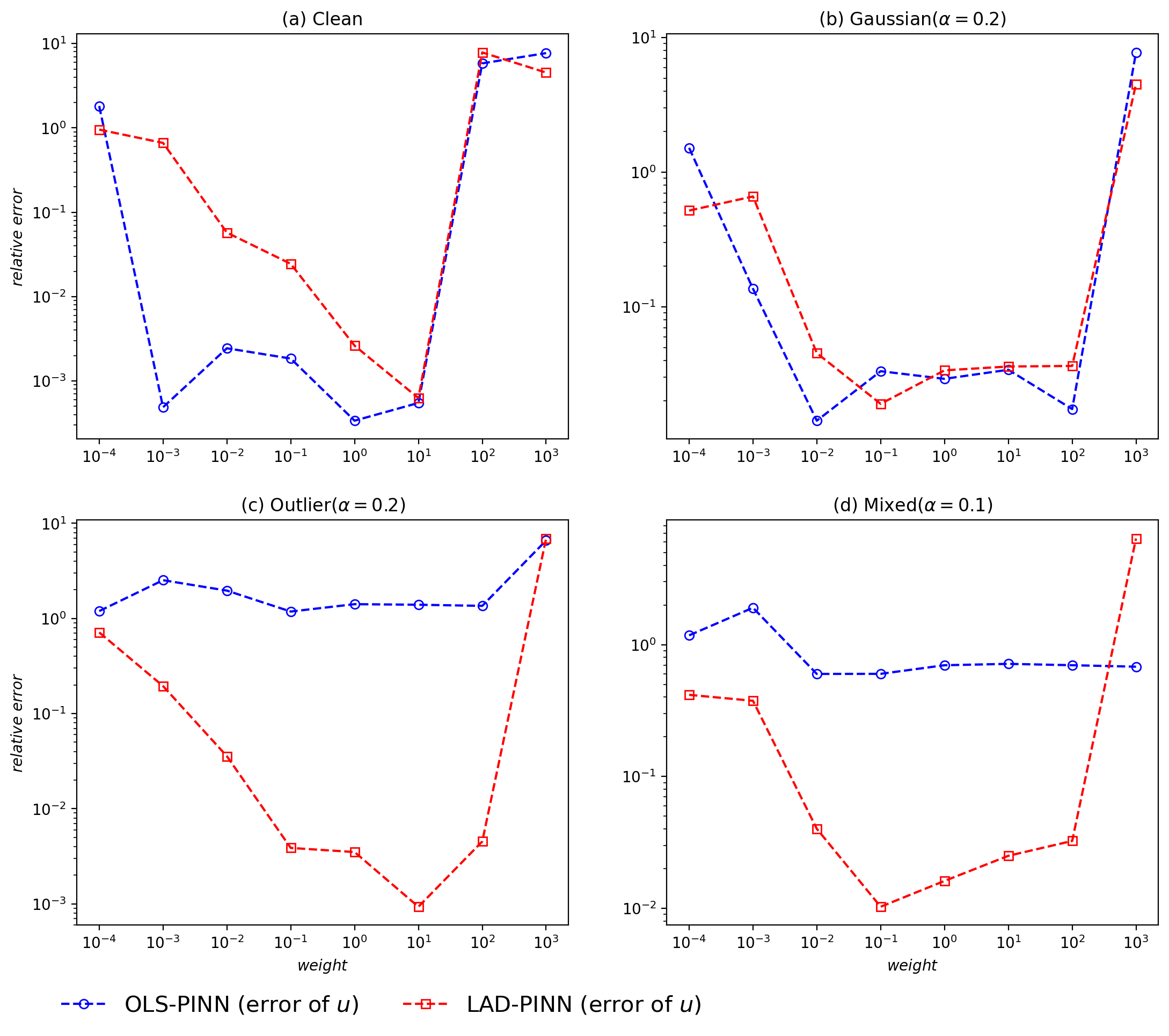}
  \caption{Effects of the weight parameter $\omega$ under different corruption types for the Poisson's equation. ($\#\mathbf{D}_u=400$, $\alpha=0.2$)}
  \label{fig:poisson_weight}
\end{figure}

The choice of weights have an impact on the results. As shown in Fig.\ref{fig:poisson_weight}, a fixed weight $\omega=1.0$ is a suitable choice for LAD-PINN with various data corruption. The weight is also suitable for OLS-PINN with clean data or Gaussian noise. However, OLS-PINN fails for all weights in the cases containing outliers.

\subsection{Steady 2D N-S equations}\label{subsec:2}
The Navier-Stokes equations describe the physics of many phenomena in science and engineering, which can be used to model weather, ocean currents, fluids in pipes and airflow around airfoils. This subsection considers a two-dimensional steady-state Navier-Stokes equations describing laminar flow, with the underlying equation given by
\begin{align}\label{eq:steadyNS}
u u_{x}+v u_{y}&=-p_{x}+u_{x x}+u_{y y}\\
u v_{x}+v v_{y}&=-p_{y}+v_{x x}+v_{y y}
\end{align}
where $u(x,y)$ denotes the x-component of the velocity field, $v(x,y)$ the y-component, and $p(x, y)$ the pressure. Incompressible solutions to the equations are constrained by the continuity equation:
\begin{align}\label{eq:incompressible}
u_x+v_y=0,
\end{align}
which describes the conservation of mass of the incompressible fluid. 

\begin{figure}[htbp]
  \centering
  \includegraphics[width=0.9\textwidth]{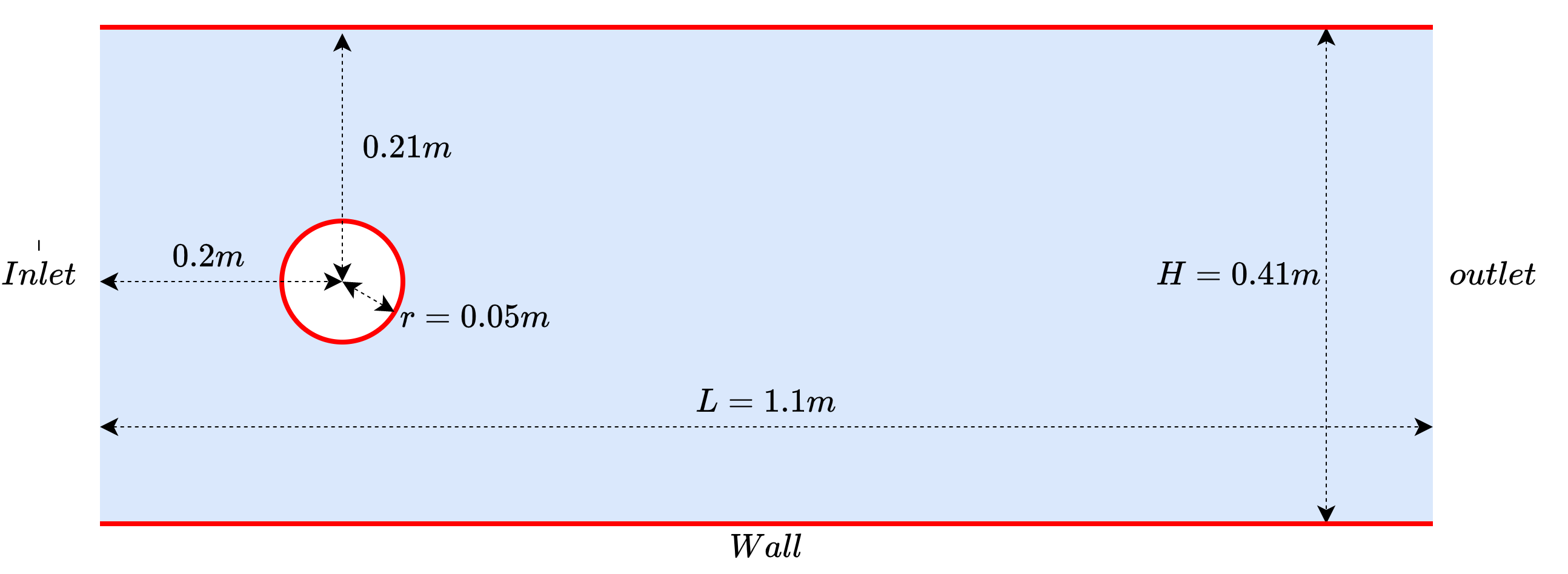}
  \caption{Problem setup of the steady N-S equation experiment. The profile of flow around a circular cylinder follows \cite{raoPhysicsinformedDeepLearning2020}. The red represents walls with the no-slip condition. We assume that the inlet boundary condition is unknown to the algorithm in the experiment.}
  \label{fig:piv_profile}
\end{figure}

The prototype problem of incompressible flow past a circular cylinder is considered. In this work, we reuse numerical example released by Rao et al.\cite{raoPhysicsinformedDeepLearning2020}. An observation in the dataset is a quadruple $(x_i,y_i,u_i,v_i)$, where the pressure is hidden from measurements. The profile is shown in Fig.\ref{fig:piv_profile}.

 The velocity vector is set to zero at all walls and the pressure is set to $p = 0$ at the outlet. The fluid density is taken as $\rho = 1kg/m^3$ and the dynamic viscosity is taken as $\mu= 2\cdot10^{-2} kg/m^3$. The velocity profile on the inlet is set as
$$
u(0, y)=4 \frac{U_{M}}{H^{2}}(H-y) y
$$
with $U_M = 1m/s$ and $H = 0.41m$. Note that the inlet profile is unknown to the algorithm, and thus we need to recover the flow fields with the incomplete boundary conditions.

To illustrate the generalization performance of the presented method, we exactly follow the network architecture proposed by Rao et al.\cite{raoPhysicsinformedDeepLearning2020}, where the Cauchy stress tensor $\sigma$ is introduced to reduce the order of derivatives in PINN.

Therefore, Eq.\eqref{eq:steadyNS} and Eq.\eqref{eq:incompressible} is equivalently transformed into the following equations:
\begin{align*}
\sigma^{11} & = -p+2\mu u_x\\
\sigma^{22}& = -p+2\mu v_y\\
\sigma^{12} &=\mu(u_y+v_x)\\
p&=-\frac{1}{2}\left(\sigma^{11}+\sigma^{22}\right)\\
\left(u u_{x}+v u_{y}\right)&=\mu\left(\sigma^{11}_x+\sigma^{12}_y\right) \\
\left(u v_{x}+v v_{y}\right)&=\mu\left(\sigma^{12}_x+\sigma^{22}_y\right)
\end{align*}
We construct a neural network with six outputs to satisfy the PDE constraints above:
\begin{align}\label{eq:piv_net}
u,v,p,\sigma^{11},\sigma^{12}, \sigma^{22}=net(x,y)
\end{align}
Samples from known boundary conditions are regarded as observations which are always clean. Combining the PDE constraints, boundaries conditions and observations, we obtain a total loss like Eq.\ref{eq:vanilla_pinn} or Eq.\ref{eq:uncon_pde}.
  
\begin{figure}[htbp]
  \centering
  \includegraphics[width=1.2\textwidth]{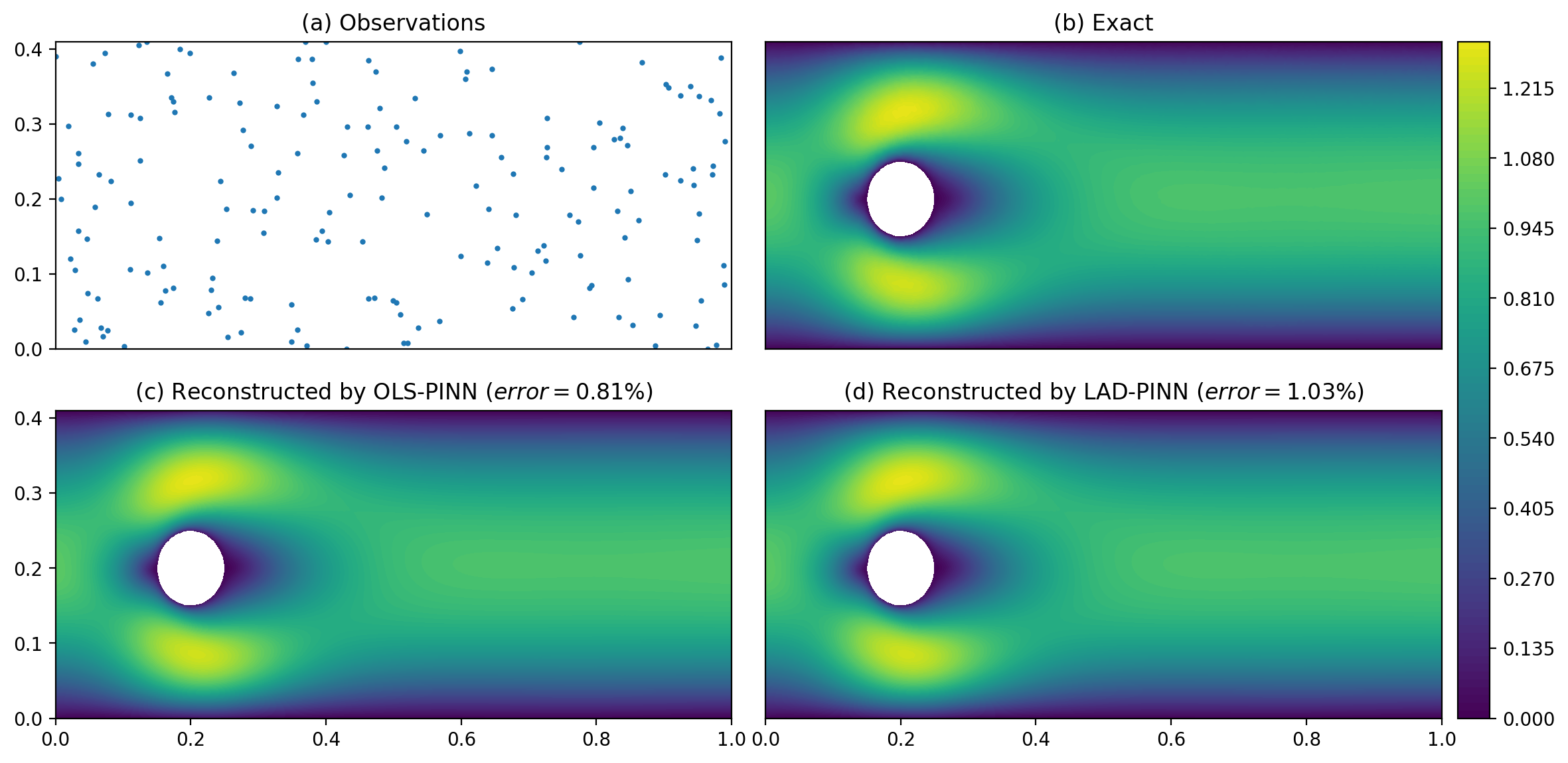}
  \caption{Magnitude of reconstructed velocity with sparse clean data. Observations are randomly located in the domain as (a) shows. Both OLS-PINN (c) and LAD-PINN (d) reach a precise reconstruction with the observations.}
  \label{fig:piv_small}
\end{figure}

\begin{figure}[htbp]
  \centering
  \includegraphics[width=1.2\textwidth]{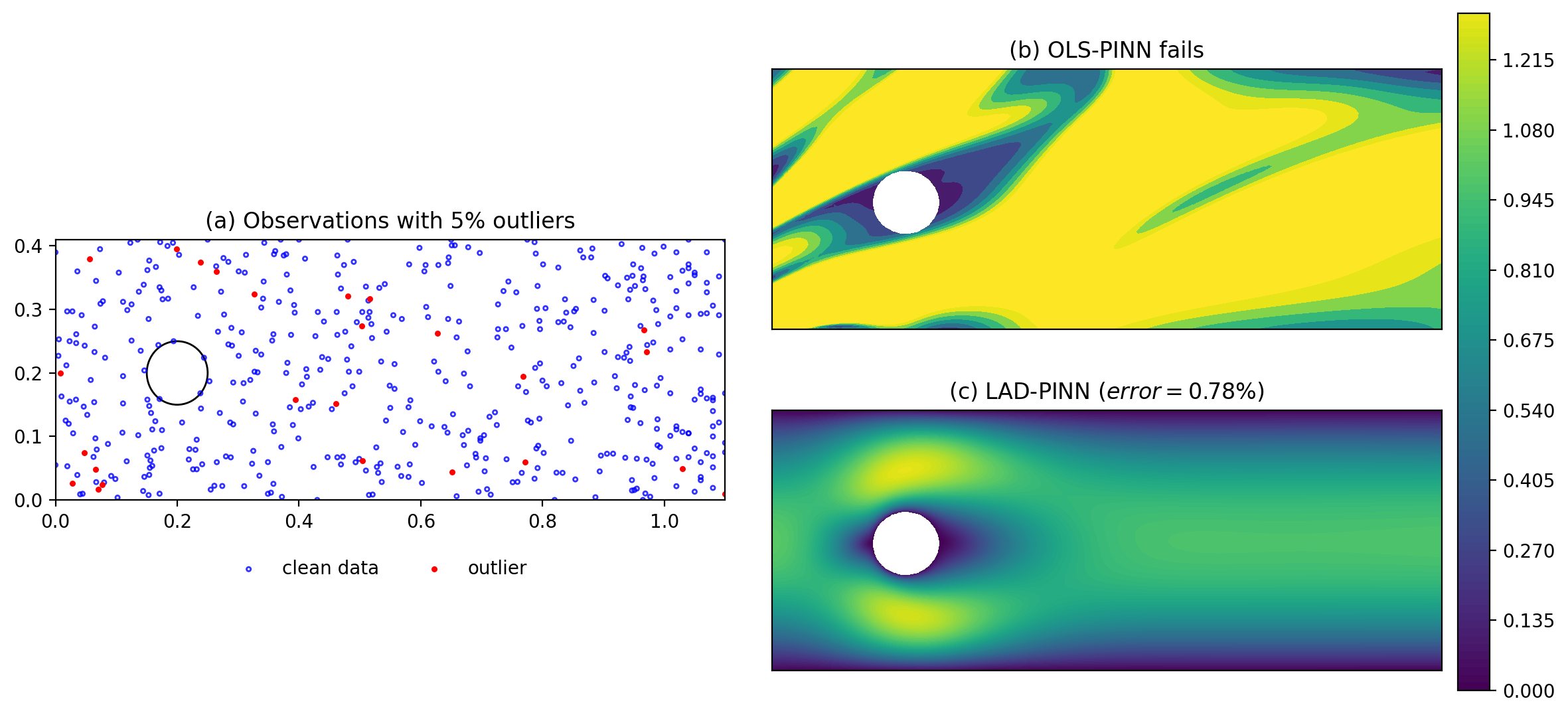}  \caption{Observations corrupted by 5\% spurious values in velocity. The  5\% outliers in the dataset are marked solid red (a). The OLS-PINN fails (b) while LAD-PINN reconstructs the velocity field even with the outliers (c).}
  \label{fig:small_bad}
\end{figure}

First we consider the noiseless case, where $u$, $v$ are recovered accurately, as shown in Fig.\ref{fig:piv_small}.
However,once a small number of outliers are added to the dataset, OLS-PINN collapses while LAD-PINN can still reconstruted the velocity field, as shown in Fig.\ref{fig:small_bad}.
For a systematic comparison, 200, 400, 600, 800, and 1000 points in the computational domain are randomly selected as observations, respectively. The results under different types of data corruption is shown in Tab.\ref{tab:size_piv}, where the corruption level $\alpha$ is fixed to $0.2$.
In this experiment, the relative errors of LAD-PINN tend to decrease as the number of observations increases. 
The results under different levels of data corruption is shown in Tab.\ref{tab:noise_piv}, where the number of observations is fixed to 1000.
The accuracy of LAD-PINN tends to decrease as the corruption level increases. 
In both experiments, except for the Gaussian case, the results of LAD-PINN are better than OLS-PINN for other data corruptions, which are consistent with the results for LAD and OLS in robust linear analysis.
\begin{table}
\begin{tabular}{ll|rrrrr}
\hline
$\#\mathbf{D}_u$ & Loss &   Gaussian &  Contaminated &    Cauchy &   Outlier &     Mixed \\
\hline
\multirow{2}{*}{200}  & LAD &     3.468 &       7.974 &   9.978 &    5.212 &   6.608 \\
     & OLS &     2.161 &      12.586 &  44.884 &  554.136 & 560.264 \\
\hline
\multirow{2}{*}{400}  & LAD &     2.957 &       3.642 &   4.724 &    2.880 &   7.485 \\
     & OLS &     2.013 &      12.129 &  43.523 &  510.407 & 525.471 \\
\hline
\multirow{2}{*}{600}  & LAD &     2.290 &       3.041 &   3.436 &    0.963 &   6.208 \\
     & OLS &     1.776 &       7.940 &  43.111 &  478.800 & 475.587 \\
\hline
\multirow{2}{*}{800}  & LAD &     2.187 &       2.805 &   3.154 &    2.307 &   2.302 \\
     & OLS &     1.553 &       5.526 &  42.126 &  475.588 & 471.305 \\
\hline
\multirow{2}{*}{1000} & LAD &     1.910 &       1.916 &   2.966 &    0.787 &   2.183 \\
     & OLS &     1.370 &       4.745 &  56.682 &  466.165 & 464.685 \\
\hline
\end{tabular}
\caption{Relative errors(\%) of reconstruct velocity fields to the steady N-S equation with different dataset sizes for $\alpha=0.2$.}
\label{tab:size_piv}
\end{table}

\begin{table}
\begin{tabular}{ll|rrrrr}
\hline
$\alpha$ & Loss     &  Gaussian &  Contaminated &    Cauchy &   Outlier &     Mixed \\
\hline
\multirow{2}{*}{0.10} & LAD &     1.031 &       1.004 &   1.453 &    0.644 &   1.361 \\
      & OLS &     0.826 &       2.260 &  23.313 &  341.776 & 341.837 \\
      \hline
\multirow{2}{*}{0.15} & LAD &     1.298 &       1.400 &   2.026 &    2.301 &   4.341 \\
      & OLS &     0.975 &       3.278 &  28.110 &  470.652 & 487.332 \\
      \hline
\multirow{2}{*}{0.20} & LAD &     1.690 &       1.647 &   2.567 &    1.975 &   5.127 \\
      & OLS &     1.209 &       4.357 &  34.447 &  598.716 & 600.624 \\
      \hline
\multirow{2}{*}{0.25} & LAD &     2.022 &       1.900 &   2.825 &    4.244 &   5.681 \\
      & OLS &     1.416 &       5.431 &  47.418 &  724.751 & 753.525 \\
      \hline
\multirow{2}{*}{0.30} & LAD &     2.324 &       2.362 &   3.294 &    5.476 &   7.311 \\
      & OLS &     1.710 &       6.917 &  53.805 &  851.395 & 850.893 \\
\hline
\end{tabular}
\caption{Relative errors(\%) of reconstructed velocity fields to the steady N-S equation with different corruption levels for $\#\mathbf{D}_u=1000$.}
\label{tab:noise_piv}
\end{table}





\subsubsection{Reveal pressure via the two stage algorithms}\label{sec:steady_pressure}
\begin{figure}[htbp]
  \centering
  \includegraphics[width=1.0\textwidth]{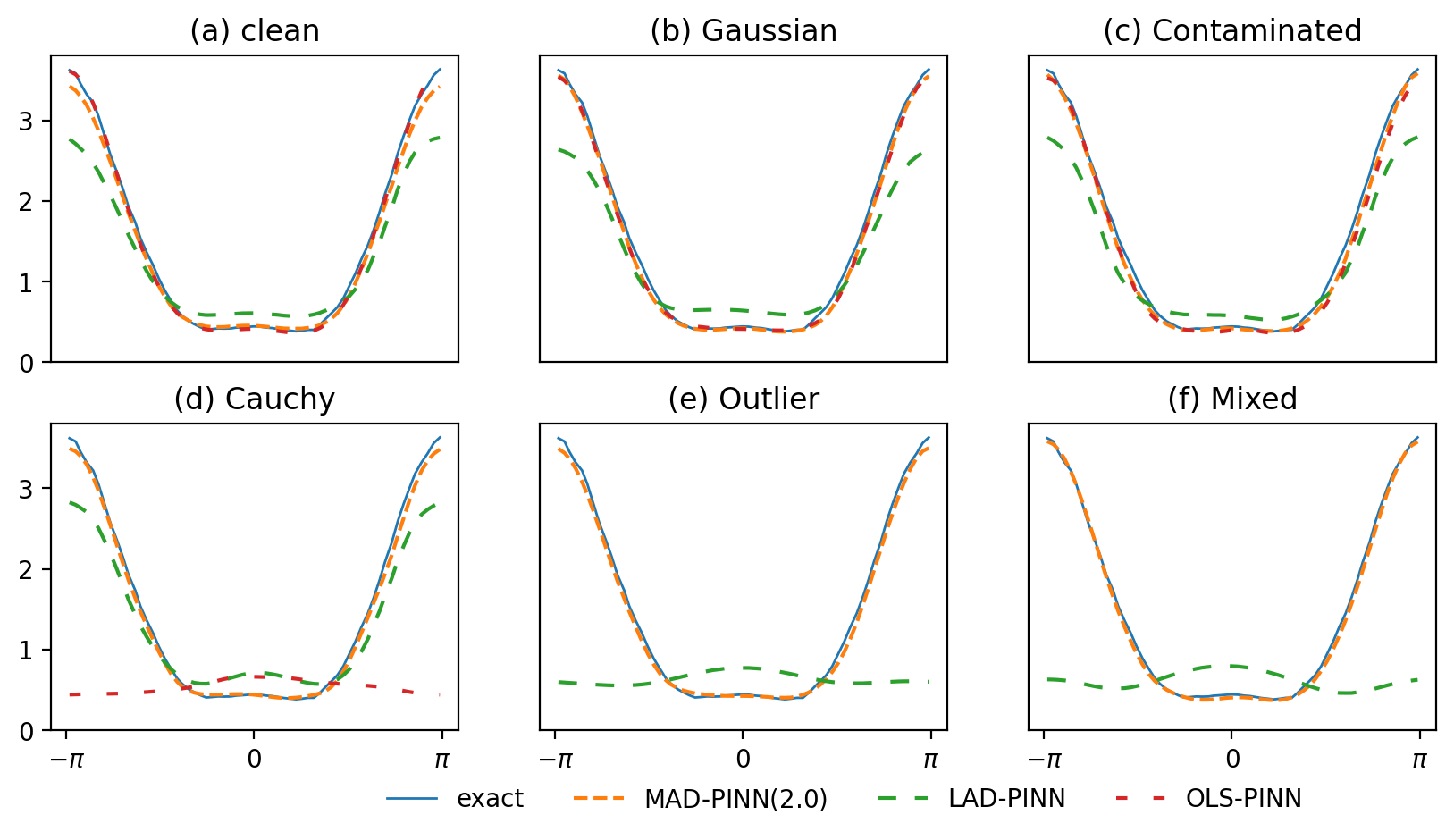}
  \caption{Reconstructed pressure around the cylinder for different corruption types. The red dash curves for OLS-PINN does not appear in (e) and (f) due to the results fall out of the displayed ranges.}
  \label{fig:piv_six_pressure}
\end{figure}

Setting $\alpha=0.2$ and $\#\mathbf{D}_u=500$, we compare the two-stage PINN framework implemented by different filter strategies and filter parameters. The result is shown in Tab.\ref{tab:piv two stage}. LAD-PINN fails to recover the pressure field with high precision. OLS-PINN collapses when highly corrupted data are involved. Except for FR(0.1), other two-stage methods recover the pressure field in relatively high accuracy ($\text{RE}<5\%$). 
The data sizes used in the second stage is also shown in the table, where MAD-PINN has small changes in the size.
The reconstructed pressure around the cylinder is shown in Fig.\ref{fig:piv_six_pressure}, where MAD-PINN(0.2) achieves an acceptable recovery.
\begin{table}
    \centering
\begin{tabular}{lc|rrrrrr}
\hline
name &  &clean &   Gaussian & Contaminated &     Cauchy &    Outlier &      Mixed \\
\hline
\multirow{2}{*}{LAD} & RE(\%) &  9.613 &   11.263 &       11.410 &   8.143 &   60.218 &   60.335 \\
         & $\#\mathbf{D}_u$ &    500 &      500 &          500 &     500 &      500 &      500 \\
         \hline
\multirow{2}{*}{OLS} & RE(\%) &  \textbf{3.086} &    4.676 &        9.805 &  76.335 &  407.450 &  587.055 \\
         & $\#\mathbf{D}_u$ &    500 &      500 &          500 &     500 &      500 &      500 \\
         \hline
\multirow{2}{*}{FR(0.1)} & RE(\%) &  3.451 &    4.285 &        5.538 &   2.757 &  958.879 &  869.857 \\
         & $\#\mathbf{\hat D}_u$ &    450 &      450 &          450 &     450 &      450 &      450 \\
         \hline
\multirow{2}{*}{FR(0.2)} & RE(\%) &  3.223 &    \textbf{4.037} &        \textbf{4.483} &   3.346 &    \textbf{3.012} &    4.407 \\
         & $\#\mathbf{\hat D}_u$ &    400 &      400 &          400 &     400 &      400 &      400 \\
         \hline
\multirow{2}{*}{FR(0.3)} & RE(\%) &  3.466 &    4.286 &        4.882 &   3.084 &    3.062 &    4.430 \\
         & $\#\mathbf{\hat D}_u$ &    350 &      350 &          350 &     350 &      350 &      350 \\
         \hline
\multirow{2}{*}{MAD(2.0)} & RE(\%) &  3.686 &    4.342 &        4.614 &   \bf{2.708} &    3.106 &    4.180 \\
         & $\#\mathbf{\hat D}_u$ &    440 &      474 &          401 &     381 &      327 &      389 \\
         \hline
\multirow{2}{*}{MAD(2.5)} & RE(\%) &  3.198 &    4.758 &        4.995 &   3.329 &    3.219 &    4.420 \\
         & $\#\mathbf{\hat D}_u$ &    460 &      493 &          414 &     399 &      340 &      391 \\
         \hline
\multirow{2}{*}{MAD(3.0)} & RE(\%) &  3.290 &    4.672 &        4.795 &   3.047 &    3.258 &    \textbf{4.119} \\
         & $\#\mathbf{\hat D}_u$ &    472 &      498 &          422 &     420 &      346 &      395 \\
\hline
\end{tabular}
    \caption{Relative errors(\%) of the reconstructed pressure $p$ in the steady N-S equation for $\#\mathbf{D}_u=500$ and $\alpha=0.2$: Bold numbers highlight the case with the best error for each corruption type.}
    \label{tab:piv two stage}
\end{table}

\begin{figure}[htbp]
  \centering
  \includegraphics[width=1.2\textwidth]{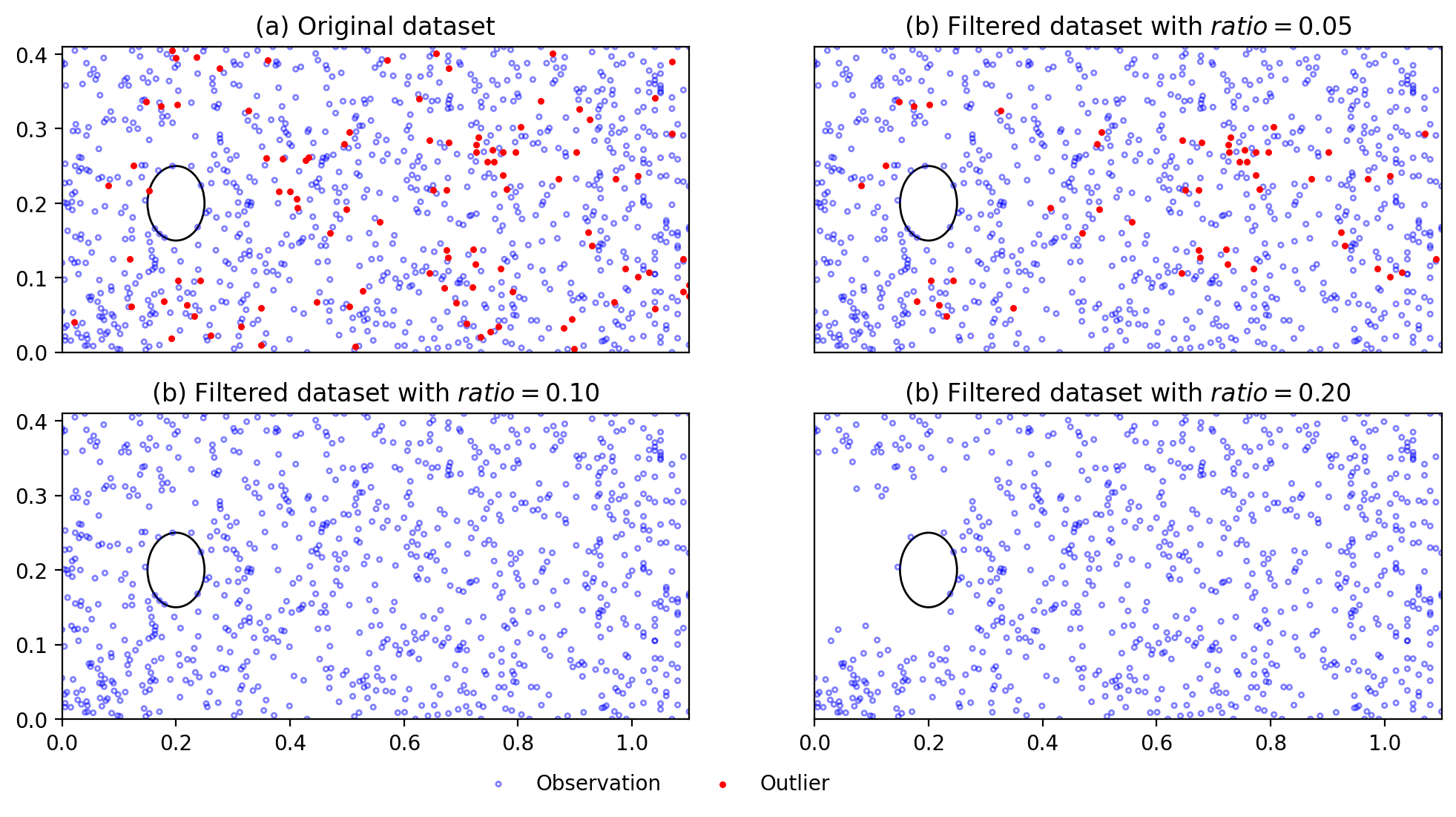}
  \caption{Different filter parameters for FR. (b) When the filter ratio is too small, the spurious values will remain in the dataset. (d) When the filter ratio is too large, all the \textit{normal} points in a small region are removed, which results in a hole without any observations near the inlet.}
  \label{fig:piv_ratio_remove}
\end{figure}

\begin{figure}[htbp]
  \centering
  \includegraphics[width=1.0\textwidth]{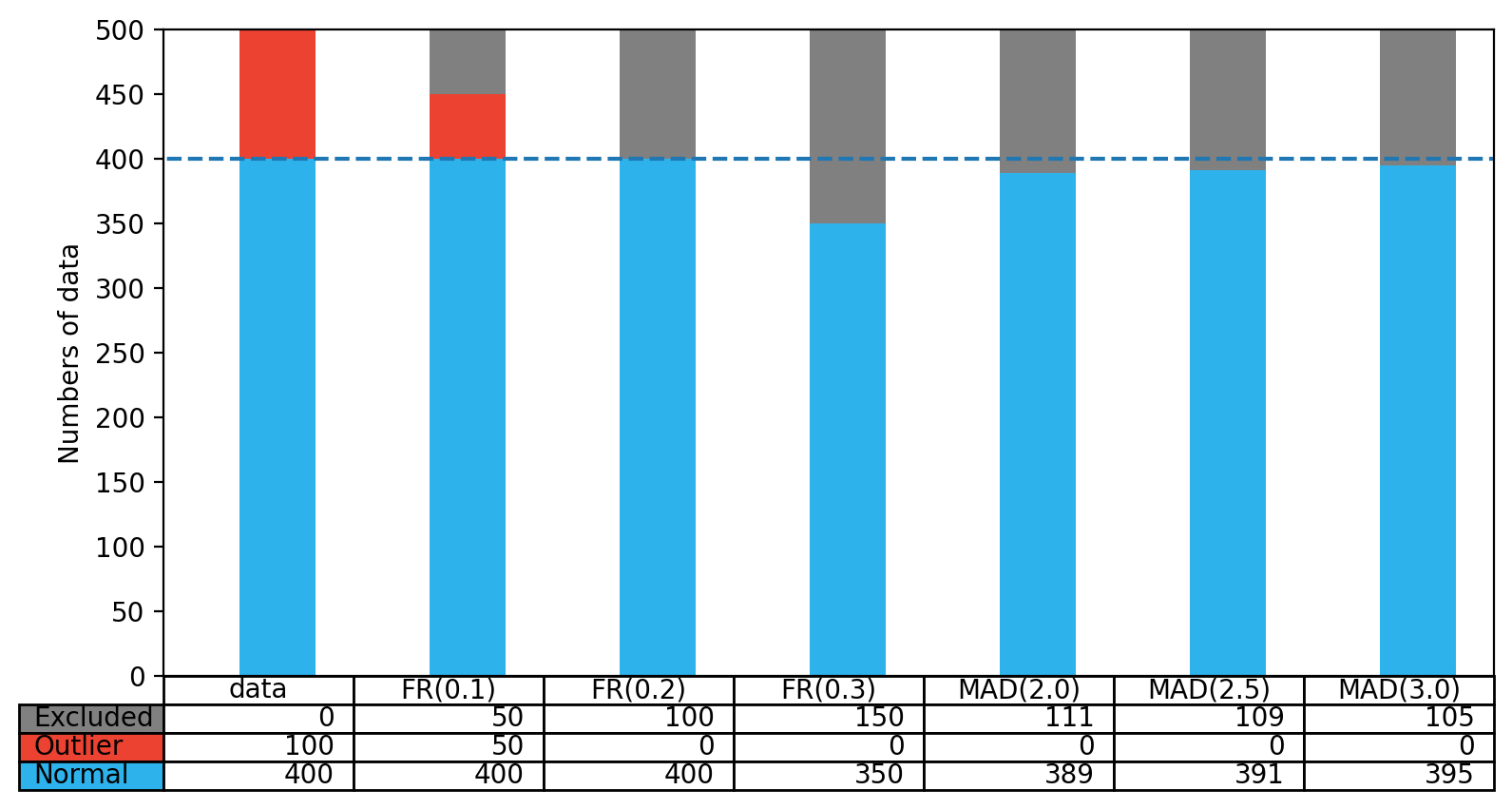}
  \caption{Composition of the observations employed in the second stage for different methods. The first column shows the observations are impaired by the Mixed corruption($\alpha=0.2$). The second column shows that FR-PINN(0.1) fails to remove all the spurious values. The last three column show that the dataset sizes remain stable for MAD-PINN with three different filter parameters.}
  \label{fig:piv_exclude_compare}
\end{figure}

In Tab.\ref{tab:piv two stage}, FR-PINN(0.1) collapses under the cases where spurious values are involved, the reason for which is as follows.
If the FR approach is used, the three situations in Fig.\ref{fig:piv_ratio_remove} may occur. When there is no accurate prior about the proportion of outliers, setting the proportion too small may cause the outliers to remain in the data set, and in Fig.\ref{fig:poisson_breaker} we have found that a small number of outliers is enough to make OLS-PINN in the second stage collapse. On the other hand, if the proportion is set too high, all the ``normal" data points in a small region are removed, creating a hole in the observations.
As shown in Figure \ref{fig:piv_exclude_compare}, MAD-PINN is not sensitive to common filter parameters(2.0, 2.5, 3.0), which does not removes too many normal points while excludes all outliers. No corruption level priors are needed, so it also indicates that the MAD-PINN may have better generalizability in practice.

\subsection{One-dimensional wave equation with unknown parameters}
\begin{figure}[htbp]
  \centering
  \includegraphics[width=1.0\textwidth]{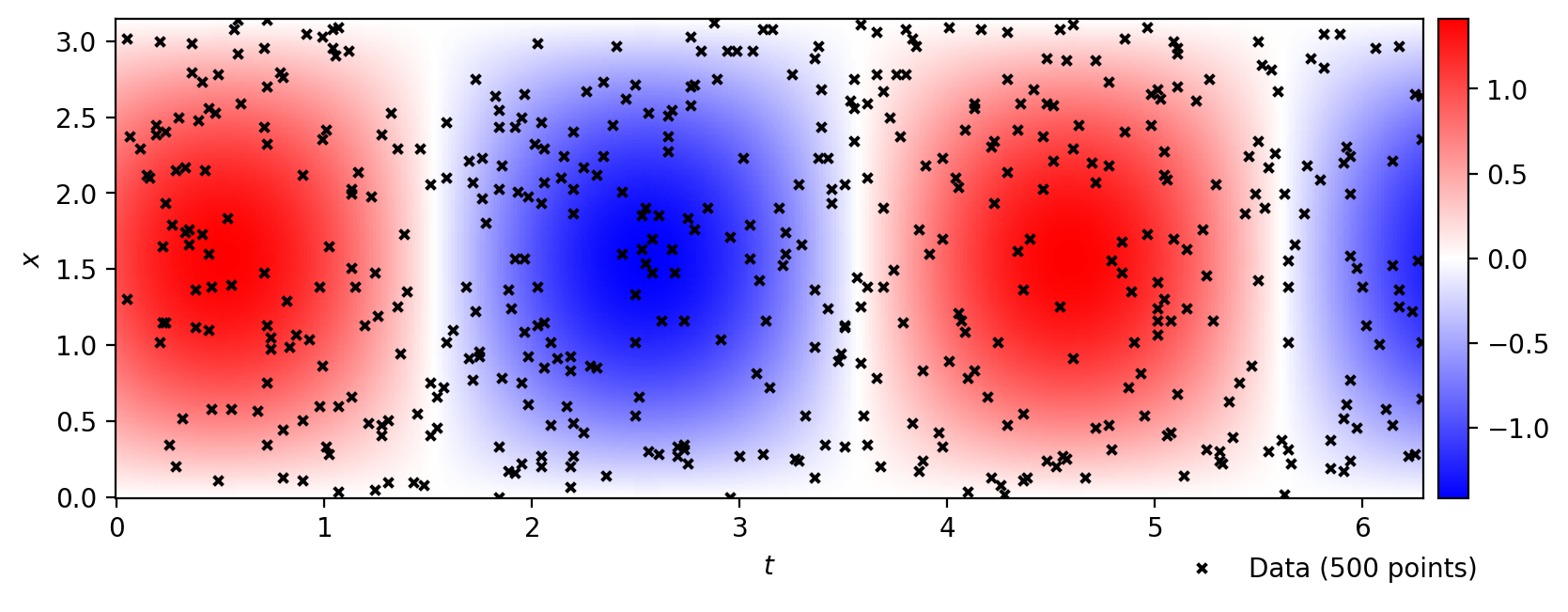}
  \caption{The ground truth of the wave equation and observations. 500 observations are randomly sampled in the interior of the domain $\Omega$.}
  \label{fig:wave_figure}
\end{figure}

\begin{figure}[htbp]
  \centering
  \includegraphics[width=1.0\textwidth]{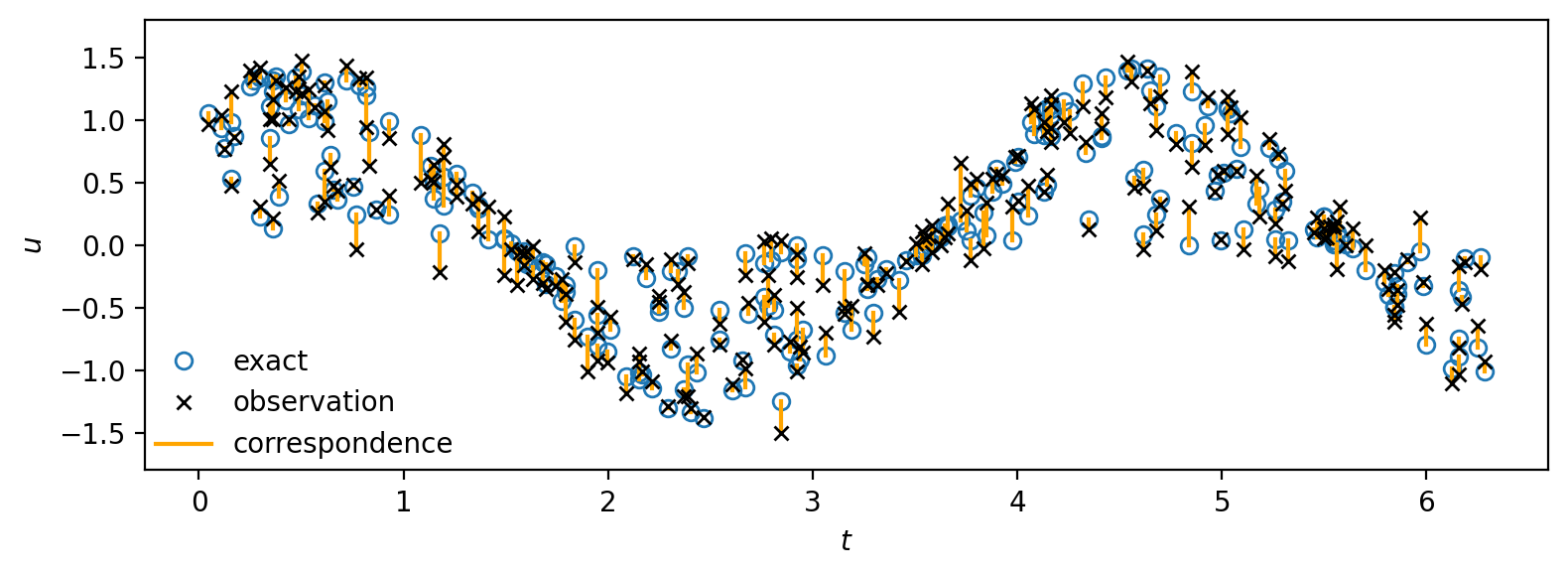}
  \caption{The exact values and the corresponding corrupted observations projected onto the $t$-$u$ plane. 250 random points are corrupted by 20\% uncorrelated Gaussian noise.}
  \label{fig:wave_noise}
\end{figure}

In this example, we consider the one-dimensional wave equation with a unknown parameter:
\begin{align}\label{eq:wave_pde}
    u_{tt}=c\cdot u_{xx}, \quad (t,x)\in\Omega=[0,2\pi]\times[0,\pi],
\end{align}
where the parameter $c =1.0$ is assumed to be unknown. The ground truth is
\begin{align}
    u=\sin x \cdot \left (\sin \sqrt{c}\cdot t + \cos \sqrt{c}\cdot t \right).
\end{align}
In this problem, we only have limided noisy data and the PDE constraint \eqref{eq:wave_pde} without any boundary or initial conditions. The Adam optimizer is applied to both the neural network parameters and the PDE parameter $c$ in the iterations.
The ground truth is illustrated in Fig.\ref{fig:wave_figure}. 
Fig.\ref{fig:wave_noise} shows 250 observations corrupted by Gaussian noise ($\alpha=0.2$).

To further scrutinize the performance of the proposed method, we have performed a systematic study with respect to the size of observation dataset, the corruption levels and the weight of the PDE loss term. First, 200, 400, 600, 800, and 1000 points in the computational domain are randomly selected as observations, respectively. The results for $u$ and $c$ under different types of data corruption is shown in Tab.\ref{tab:wave_error_u_size} and Tab.\ref{tab:wave_error_c_size}, where the corruption level $\alpha$ is fixed to $0.1$. The results under different levels of data corruption is shown in Tab.\ref{tab:wave_error_u} and Tab.\ref{tab:wave_error_c}, where the number of observations is fixed to 1000. In the clean cases, OLS-PINN performs better than LAD-PINN. In the Gaussian cases, OLS-PINN is comparable with LAD-PINN. For the remaining highly corrupted cases, OLS-PINN usually has large errors or even crashes. The results of LAD-PINN are more accurate and very stable.

\begin{table}
\begin{tabular}{ll|rrrrrr}
\hline
$\#\mathbf{D}_u$ & loss       &  Clean &  Gaussian &  Contaminated &  Cauchy &  Outlier &   Mixed \\
\hline
\hline
\multirow{2}{*}{200}  & LAD &  2.492 &     5.402 &          6.084 &  14.699 &    3.169 &   7.987 \\
     & OLS &  0.495 &    11.244 &        255.549 & 118.851 &  497.638 & 554.716 \\\hline
\multirow{2}{*}{400}  & LAD &  0.707 &     3.341 &          4.252 &   4.442 &    1.622 &   4.353 \\
     & OLS &  0.362 &     2.384 &        261.212 &  82.617 &  368.784 & 377.041 \\\hline
\multirow{2}{*}{600}  & LAD &  0.429 &     2.684 &          3.055 &   3.936 &    0.600 &   3.211 \\
     & OLS &  0.109 &     2.104 &        174.771 & 323.152 &  274.044 & 273.428 \\\hline
\multirow{2}{*}{800}  & LAD &  0.692 &     2.211 &          2.496 &   2.853 &    0.504 &   2.953 \\
     & OLS &  0.191 &     1.967 &        169.544 & 317.465 &  211.707 & 213.715 \\\hline
\multirow{2}{*}{1000} & LAD &  0.565 &     2.086 &          2.456 &   3.090 &    1.766 &   2.595 \\
     & OLS &  0.127 &     1.513 &        135.489 & 251.246 &  314.845 & 303.172 \\
\hline
\end{tabular}\caption{Relative errors(\%) of the solution $u$ to the wave equation with different dataset sizes for $\alpha=0.1$.}
\label{tab:wave_error_u_size}
\end{table}

\begin{table}
\begin{tabular}{ll|rrrrrr}
\hline
$\#\mathbf{D}_u$ & loss       &  Clean &  Gaussian &  Contaminated &  Cauchy &  outlier &  mixed \\
\hline\hline
\multirow{2}{*}{200}  & LAD &  4.362 &     6.615 &          4.842 &  14.399 &    9.231 &  9.251 \\
     & OLS &  2.517 &     2.625 &         52.035 &  80.788 &   97.263 & 93.979 \\\hline
\multirow{2}{*}{400}  & LAD &  3.087 &     2.146 &          1.161 &   3.920 &    5.217 &  3.783 \\
     & OLS &  2.561 &     0.613 &         85.672 &   8.559 &   69.222 & 71.760 \\\hline
\multirow{2}{*}{600}  & LAD &  2.962 &     2.672 &          2.933 &   2.431 &    2.632 &  3.040 \\
     & OLS &  2.526 &     2.898 &         53.954 &  42.297 &   52.241 & 52.404 \\\hline
\multirow{2}{*}{800}  & LAD &  2.604 &     2.029 &          3.199 &   2.344 &    2.359 &  3.128 \\
     & OLS &  2.546 &     2.602 &         57.855 &  69.769 &   33.706 & 32.523 \\\hline
\multirow{2}{*}{1000} & LAD &  2.867 &     2.303 &          2.779 &   1.804 &    3.613 &  3.481 \\
     & OLS &  2.550 &     2.814 &         37.539 &  35.793 &   98.799 & 98.314 \\
\hline
\end{tabular}
\caption{Relative errors(\%) of the recovered parameter $c$ in the wave equation with different dataset sizes for $\alpha=0.1$.}
\label{tab:wave_error_c_size}
\end{table}

\begin{table}
\begin{tabular}{ll|rrrrr}
\hline
$\alpha$      & loss       &  Gaussian &  Contaminated &  Cauchy &  Outlier &  Mixed \\
\hline\hline
\multirow{2}{*}{0.05} & LAD &     2.446 &          2.322 &   2.073 &    3.187 &  3.304 \\
      & OLS &     2.644 &          0.123 &   7.411 &   94.105 & 94.711 \\\hline
\multirow{2}{*}{0.10} & LAD &     2.303 &          1.848 &   1.804 &    3.613 &  3.481 \\
      & OLS &     2.814 &          7.379 &  35.793 &   98.799 & 98.314 \\\hline
\multirow{2}{*}{0.15} & LAD &     2.609 &          2.274 &   1.199 &    2.676 &  2.682 \\
      & OLS &     2.672 &          8.371 &  40.515 &   93.946 & 94.022 \\\hline
\multirow{2}{*}{0.20} & LAD &     2.730 &          2.257 &   0.593 &    2.999 &  3.632 \\
      & OLS &     2.932 &         15.652 &  38.493 &   94.681 & 93.448 \\
\hline
\end{tabular}
\caption{Relative errors(\%) of the solution $u$ to the wave equation with different corruption levels for $\#\mathbf{D}_u=1000$.}
\label{tab:wave_error_u}
\end{table}

\begin{table}
\begin{tabular}{ll|rrrrr}
\hline
$\alpha$ & loss      &  Gaussian &  Contaminated &  Cauchy &  Outlier &   Mixed \\
\hline\hline
\multirow{2}{*}{0.05} & LAD &     1.239 &          1.641 &   2.010 &    1.052 &   1.368 \\
      & OLS &     0.899 &          3.472 & 117.890 &  200.846 & 205.928 \\\hline
\multirow{2}{*}{0.10} & LAD &     2.086 &          2.438 &   3.090 &    1.766 &   2.595 \\
      & OLS &     1.513 &         11.431 & 251.246 &  314.845 & 303.172 \\\hline
\multirow{2}{*}{0.15} & LAD &     2.965 &          3.874 &   3.464 &    0.500 &   4.569 \\
      & OLS &     2.169 &         11.030 & 385.941 &  367.705 & 362.870 \\\hline
\multirow{2}{*}{0.20} & LAD &     3.673 &          5.121 &   4.748 &    0.667 &   8.476 \\
      & OLS &     2.905 &         16.749 & 470.102 &  474.444 & 386.001 \\
\hline
\end{tabular}
\caption{Relative errors(\%) of the recovered parameter $c$ in the wave equation with different corruption levels for $\#\mathbf{D}_u=1000$.}
\label{tab:wave_error_c}
\end{table}

\begin{figure}[htbp]
  \centering
  \includegraphics[width=1.0\textwidth]{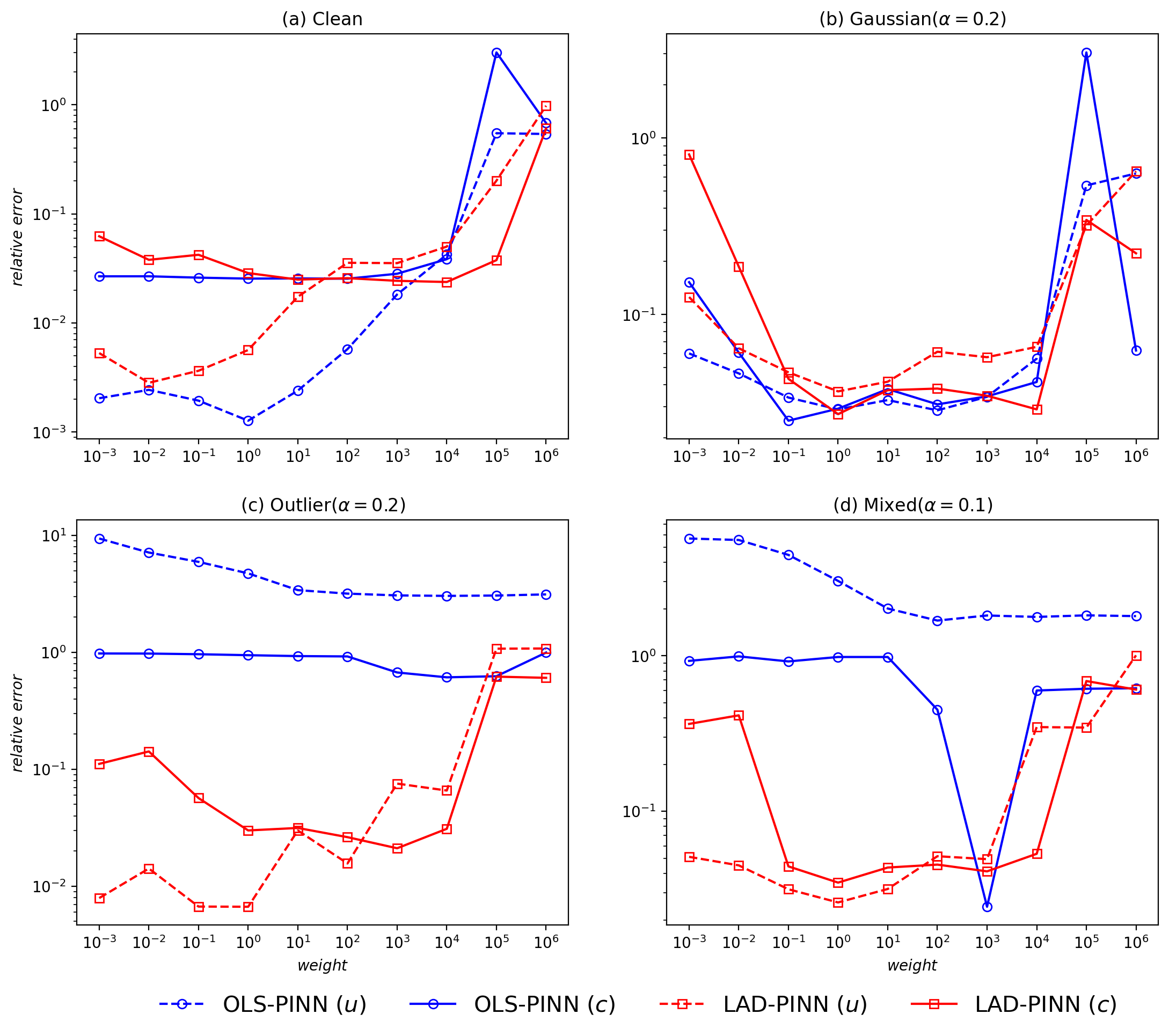}
  \caption{Effects of the weight parameter $\omega$ under different corruption types for the wave equation. ($\#\mathbf{D}_u=1000$, $\alpha=0.2$)}
  \label{fig:wave_weight}
\end{figure}

Finally, let us investigate the impacts of different weights. Fix the dataset size to be 1000, we vary the weight from $10^{-3}$ to $10^6$. As is shown in Fig.\ref{fig:wave_weight}, for the clean and Gaussian cases, both LAD-PINN and OLS-PINN with $\omega=1.0$ reconstruct the physical field and recover the parameters relatively well. For all cases, LAD-PINN with $\omega=1.0$ could achieve acceptable small errors. Therefore, choosing $\omega=1.0$ might be appropriate in this subsection.

\subsection{Unsteady 2D N-S equations with unknown parameters}
In the last experiment, unsteady N-S equations are considered, the problem setup of which is referred to \cite{raissiPhysicsinformedNeuralNetworks2019}. The N–S equations are explicitly given by
\begin{align}
    u_t+\lambda_{1}\left(u u_{x}+v u_{y}\right)&=-p_{x}+\lambda_{2}\left(u_{x x}+u_{y y}\right) \\
    v_t+\lambda_{1}\left(u v_{x}+v v_{y}\right)&=-p_{y}+\lambda_{2}\left(v_{x x}+v_{y y}\right)
\end{align}
where $\lambda_1$ and $\lambda_2$ are two unknown parameters to be recovered. We make the assumption that
\begin{align*}
    u=\psi_{y}, \quad v=-\psi_{x}
\end{align*}
for some stream function $\psi(x,y)$. Under this assumption, the continuity equation will be automatically satisfied. The following architecture is used in this example,
\begin{align}\label{eq:unsteady_net} 
    \psi, p = net(t,x,y,\lambda_1,\lambda_2).
\end{align}
Note that unlike the neural network architecture used in Eq.\eqref{eq:piv_net}, 
network \eqref{eq:unsteady_net} has only two equation constraints, while the former has six equation constraints.
The advantage of network \eqref{eq:piv_net} is that it only requires solving the first-order derivatives while network \eqref{eq:unsteady_net} requires solving the third-order derivatives.
Fewer loss terms or lower order derivatives may both make the training of PINN easier. 
The two architectures are not compared in this paper, which are only employed to illustrate the generalization of the proposed method.

\begin{figure}[htbp]
  \centering
  \includegraphics[width=1.0\textwidth]{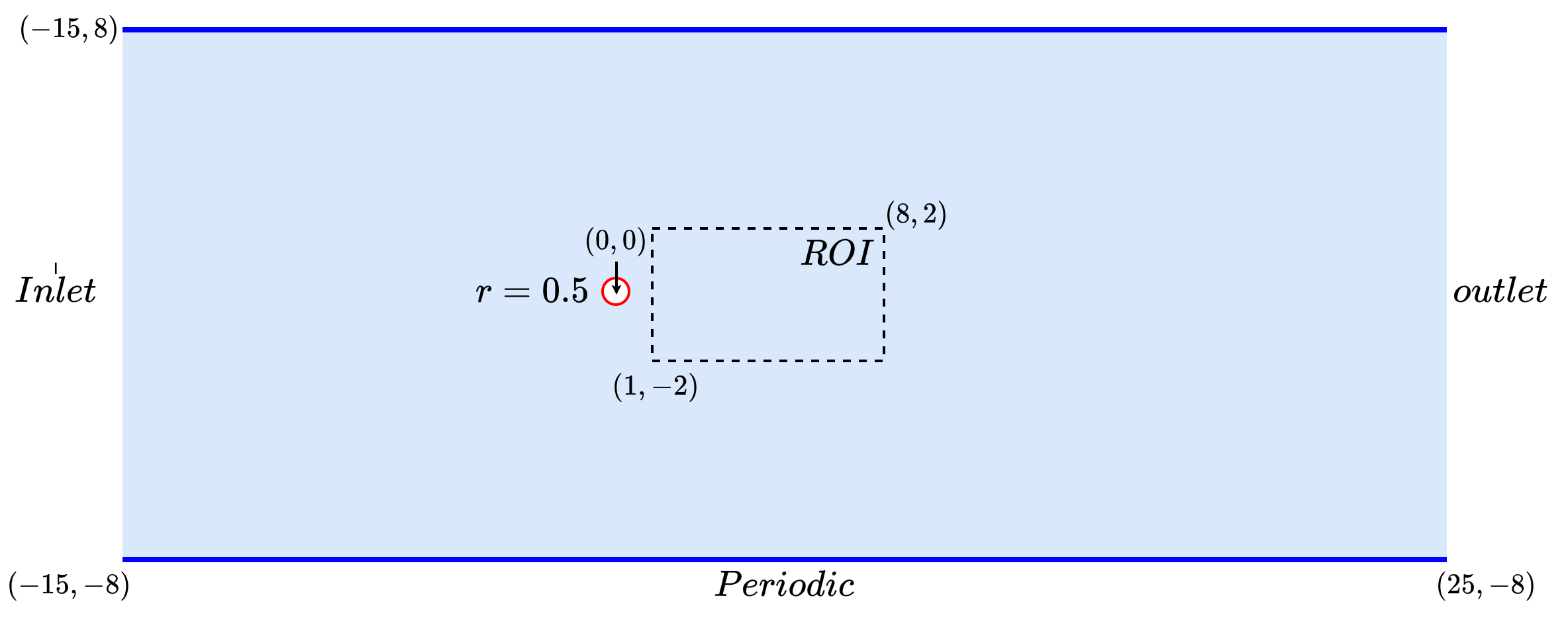}
  \caption{Problem setup of the unsteady N-S equation experiment. 
  The spatio-temporal training data are sampled from the rectangular ROI behind the cylinder which is marked red. Refer to \cite{raissiPhysicsinformedNeuralNetworks2019} for a detailed profile.}
  \label{fig:unsteady-struct}
\end{figure}

Consider the rectangular region of interest (ROI) as shown in Fig.\ref{fig:unsteady-struct}.

We set $\lambda_1=1.0$ and $\lambda_2=0.01$. An observation has the form of quintuple $(t_i,x_i,y_i,u_i,v_i)$, where pressure is hidden from observing. The data corruption setting is similar to Subsection \ref{subsec:2}. 

For the recovery of $\lambda_1$ (Tab.\ref{tab:nslambda1alpha}), we first consider fix $\#\mathbf{D}_u=5000$. OLS-PINN performs much more accurate than LAD-PINN in four Gaussian cases and Contaminated cases under two lower corruption levels. When the data becomes more highly corrupted, LAD-PINN stays stable while OLS-PINN begins to crash. In Tab.\ref{tab:nslambda1N}, we fix $\alpha=0.05$, which is small. Then OLS-PINN performs better than LAD-PINN in all Gaussian and Contaminated cases but worse than LAD-PINN in other highly corrupted cases.

For the recovery of $\lambda_2$ (Tab.\ref{tab:nslambda2alpha} and Tab.\ref{tab:nslambda2N}). OLS-PINN performs similarly to its performance in the recovery of $\lambda_1$. OLS-PINN  recovers the parameter in less corrupted cases (Gaussian and Contaminated), but crashes in the remaining highly corrupted cases. Instead, LAD-PINN is much stable for the recovery of $\lambda$ in all cases.

\begin{table}
\begin{tabular}{ll|rrrrr}
\hline
$\alpha$ &   loss     &  Gaussian &  Contaminated &    Cauchy &   Outlier &     Mixed \\
\hline\hline
\multirow{2}{*}{0.01} & LAD &     0.536 &       0.634 &   0.901 &    0.694 &   0.435 \\
      & OLS &     0.162 &       0.276 &   2.459 &  100.109 &  99.961 \\\hline
\multirow{2}{*}{0.05} & LAD &     0.651 &       0.588 &   0.804 &    1.392 &   1.009 \\
      & OLS &     0.050 &       0.058 &  99.749 &  104.030 & 106.178 \\\hline
\multirow{2}{*}{0.10} & LAD &     0.664 &       0.174 &   0.917 &    0.672 &   1.426 \\
      & OLS &     0.014 &       0.702 &  99.943 &   99.621 &  99.548 \\\hline
\multirow{2}{*}{0.20} & LAD &     0.857 &       1.038 &   1.301 &    1.690 &   4.062 \\
      & OLS &     0.045 &       6.847 &  98.114 &  101.828 & 101.638 \\
\hline
\end{tabular}
\caption{Relative errors(\%) of the recovered parameter $\lambda_1$ in the unsteady N-S equation with different corruption levels for $\#\mathbf{D}_u=5000$.}\label{tab:nslambda1alpha}
\end{table}

\begin{table}
\begin{tabular}{ll|rrrrr}
\hline
   $\alpha$   &     loss   &  Gaussian &  Contaminated &  Cauchy &  Outlier &   Mixed \\
\hline\hline
\multirow{2}{*}{0.01} & LAD &     7.475 &       5.252 &   7.314 &    1.404 &   5.252 \\
      & OLS &     8.542 &      10.755 &  67.498 &  100.047 & 100.062 \\\hline
\multirow{2}{*}{0.05} & LAD &     3.613 &       9.728 &   9.294 &   12.161 &   2.975 \\
      & OLS &     6.871 &       9.172 & 100.065 &   99.976 &  99.865 \\\hline
\multirow{2}{*}{0.10} & LAD &     1.557 &       5.987 &   9.090 &    3.438 &   0.894 \\
      & OLS &     3.740 &      17.366 & 100.166 &   99.961 &  99.997 \\\hline
\multirow{2}{*}{0.20} & LAD &    10.564 &      12.006 &   9.069 &    5.681 &  12.439 \\
      & OLS &     0.989 &      80.157 &  18.411 &   99.936 & 100.007 \\
\hline
\end{tabular}
\caption{Relative errors(\%) of the recovered parameter $\lambda_2$ in the unsteady N-S equation with different corruption levels for $\#\mathbf{D}_u=5000$.}\label{tab:nslambda2alpha}
\end{table}

\begin{table}
\begin{tabular}{ll|rrrrr}
\hline
N & loss &    Gaussian &  Contaminated &  Cauchy &  Outlier &   Mixed \\
\hline\hline
\multirow{2}{*}{2000} & LAD &     0.377 &       1.263 &   1.123 &    0.831 &   1.009 \\
     & OLS &     0.048 &       0.390 & 100.431 &   95.808 &  92.520 \\\hline
\multirow{2}{*}{4000} & LAD &     0.844 &       0.921 &   1.050 &    1.094 &   0.760 \\
     & OLS &     0.003 &       0.160 &  99.264 &  105.737 & 101.559 \\\hline
\multirow{2}{*}{6000} & LAD &     0.578 &       0.813 &   0.897 &    0.801 &   0.609 \\
     & OLS &     0.105 &       0.342 &  88.314 &   99.692 &  99.869 \\\hline
\multirow{2}{*}{8000} & LAD &     0.500 &       0.495 &   0.452 &    0.651 &   0.592 \\
     & OLS &     0.152 &       0.265 &  66.310 &  101.267 & 101.817 \\
\hline
\end{tabular}
\caption{Relative errors(\%) of the recovered parameter $\lambda_1$ in the unsteady N-S equation with different dataset sizes for $\alpha=0.05$.}\label{tab:nslambda1N}
\end{table}

\begin{table}
\begin{tabular}{ll|rrrrr}
\hline
N & loss  &    Gaussian &  Contaminated &  Cauchy &  Outlier &   Mixed \\
\hline\hline
\multirow{2}{*}{2000} & LAD &     1.488 &       0.656 &   6.401 &    1.557 &   0.514 \\
     & OLS &     4.464 &       5.262 &  98.688 &  100.210 & 100.125 \\\hline
\multirow{2}{*}{4000} & LAD &     3.068 &       7.197 &   0.827 &    2.848 &   3.895 \\
     & OLS &     7.422 &       7.094 &  99.636 &  100.063 &  99.995 \\\hline
\multirow{2}{*}{6000} & LAD &     3.784 &       5.889 &   6.806 &    2.006 &   3.549 \\
     & OLS &     7.547 &       4.098 &  99.480 &  100.028 & 100.000 \\\hline
\multirow{2}{*}{8000} & LAD &     4.451 &       5.455 &   0.326 &    1.479 &   0.969 \\
     & OLS &     8.108 &       5.003 & 105.482 &   99.929 & 100.047 \\
\hline
\end{tabular}
\caption{Relative errors(\%) of the recovered parameter $\lambda_2$ in the unsteady N-S equation with different dataset sizes for $\alpha=0.05$.}\label{tab:nslambda2N}
\end{table}

\subsubsection{Reveal pressure via the two stage algorithms}\label{sec:unsteady_pressure}
Since the dynamic pressure field is hidden, we employ the two-stage framework to reveal the field from corrupted velocity observations. For a given moment $t_p$, the reconstructed pressure and the reference will differ by a constant. 
We define the constant by minimizing the squared norms:
\begin{align}
    \hat c=\arg\min_c \|p_{ref}-\hat p+c\|_2^2
\end{align}
Due to the first-order optimality condition, we have
\begin{align}
\hat c:=\frac{\int_\Omega \left(\hat p-p_{ref}\right)dxdy}{\int_\Omega dxdy}.
\end{align}
Then we define the relative $l_2$ error to measure the accuracy of the reconstructed pressure as follows:
\begin{align}
    rel_p:=\frac{\|p_{ref}-\hat p+\hat c\|_2}{\|p_{ref}\|_2}.
\end{align}

\begin{figure}[htbp]
  \centering
  \includegraphics[width=1.0\textwidth]{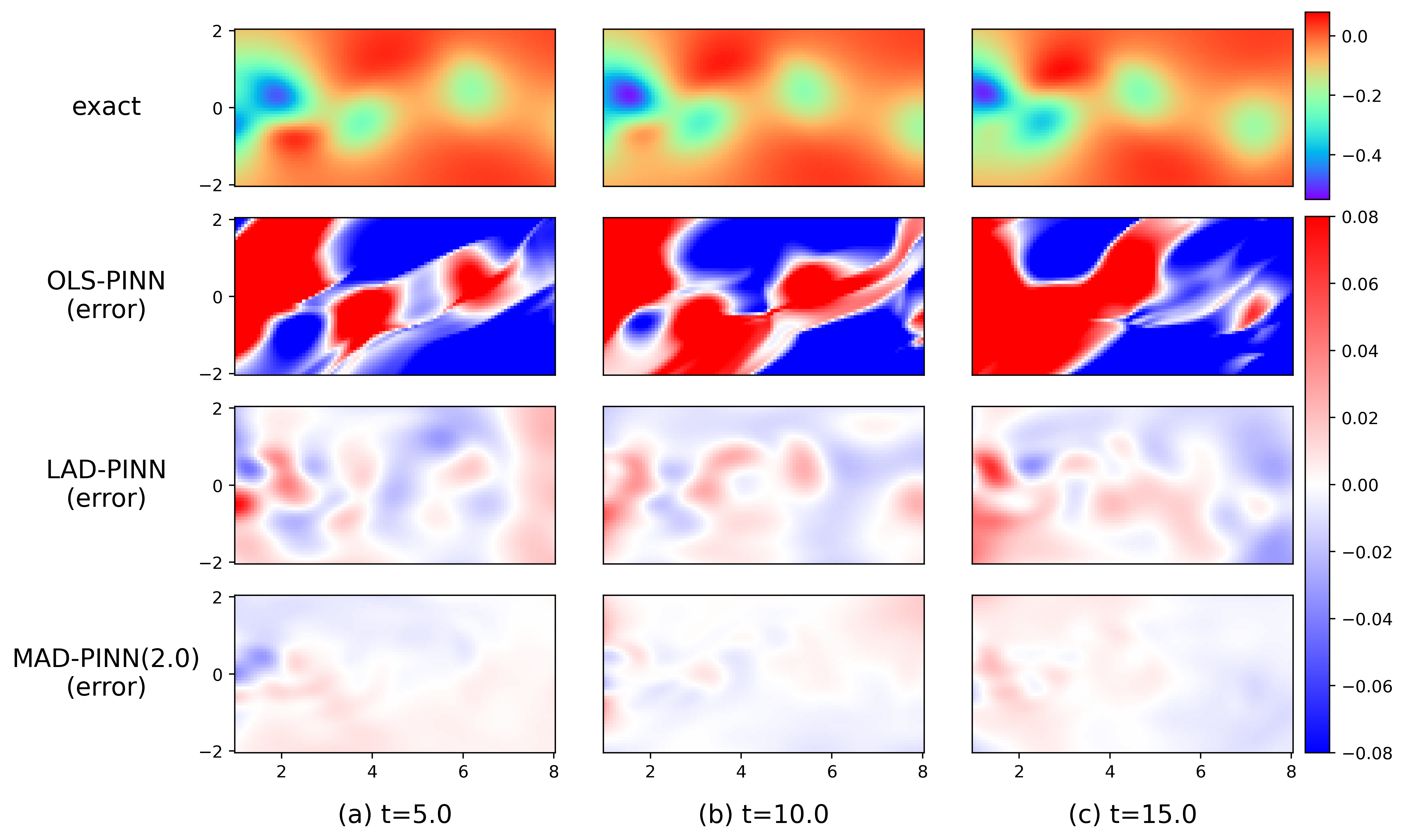}
  \caption{The snapshots of the ground truth pressure (the first row) and reconstruction errors (the last three rows) for the unsteady N-S equation. The magnitudes of OLS-PINN's error fall out of the colorbar.}
  \label{fig:two_stage_snapshot}
\end{figure}

An snapshot at $t = 10.0$ is considered and the results are shown in Table \ref{tab:re_unsteady_two_stage}. Except for the clean and Gaussian cases, the single-stage algorithm OLS-PINN cannot achieve an acceptable results, where the accuracy is very low or even explodes in all other cases. Single-stage LAD-PINN is usually less accurate, although it can reconstruct the pressure field under all kinds of data corruption. Two-stage algorithms can achieve more accurate results. Fig.\ref{fig:two_stage_snapshot} shows the visualization results of the pressure at $t=5.0, 10.0, 15.0$, and it can be seen that MAD-PINN has the smallest error among the three. This example demonstrates the ability of the two-stage methods to reveal hidden physics accurately.

\begin{table}
\begin{tabular}{lc|rrrrrr}
\hline
Method &           & Clean &   Gaussian & Contaminated &     Cauchy &    Outlier &      Mixed \\
\hline\hline
\multirow{2}{*}{LAD}     & RE(\%) &  6.259 &    7.018 &        9.463 &    9.479 &  10.328 &    8.126 \\
         & $\#\mathbf{D}_u$ &   5000 &     5000 &         5000 &     5000 &    5000 &     5000 \\\hline
\multirow{2}{*}{OLS}    & RE(\%) &  3.336 &    \textbf{3.287} &       10.294 &  108.423 &  76.746 &  103.908 \\
         & $\#\mathbf{D}_u$ &   5000 &     5000 &         5000 &     5000 &    5000 &     5000 \\\hline
\multirow{2}{*}{FR(0.1)} & RE(\%) &  2.484 &    4.053 &        8.933 &    7.576 &   \textbf{1.773} &    \textbf{3.359} \\
         & $\#\mathbf{\hat D}_u$ &   4500 &     4500 &         4500 &     4500 &    4500 &     4500 \\\hline
\multirow{2}{*}{FR(0.2)} & RE(\%) &  3.181 &    4.618 &        \textbf{4.248} &    6.419 &   2.073 &    4.136 \\
         & $\#\mathbf{\hat D}_u$ &   4000 &     4000 &         4000 &     4000 &    4000 &     4000 \\\hline
\multirow{2}{*}{FR(0.3)} & RE(\%) &  3.131 &    4.717 &        4.823 &    \textbf{5.528} &   2.423 &    4.496 \\
         & $\#\mathbf{\hat D}_u$ &   3500 &     3500 &         3500 &     3500 &    3500 &     3500 \\\hline
\multirow{2}{*}{MAD(2.0)} & RE(\%) &  2.885 &    3.814 &        4.592 &    6.618 &   2.259 &    3.589 \\
         & $\#\mathbf{\hat D}_u$ &   4401 &     4936 &         4069 &     4015 &    4035 &     4453 \\\hline
\multirow{2}{*}{MAD(2.5)} & RE(\%) &  2.317 &    3.751 &        4.977 &    9.473 &   2.132 &    3.492 \\
         & $\#\mathbf{\hat D}_u$ &   4597 &     4971 &         4123 &     4200 &    4176 &     4474 \\\hline
\multirow{2}{*}{MAD(3.0)} & RE(\%) &  \textbf{2.116} &    3.542 &        4.822 &    8.416 &   2.164 &    3.669 \\
         & $\#\mathbf{\hat D}_u$ &   4709 &     4981 &         4190 &     4336 &    4268 &     4477 \\
\hline
\end{tabular}
\caption{Relative errors(\%) of the reconstructed pressure $p$ in the unsteady N-S equation at $t=10.0$ for $\#\mathbf{D}_u=5000$ and $\alpha=0.1$: Bold numbers highlight the case with the best error for each corruption type.}
\label{tab:re_unsteady_two_stage}
\end{table}
\section{Discussion and Conclusion}

In order to apply PINN to a wide range of practical tasks, it is necessary to consider the cases involving anomalous data. Instead of employing data cleaning as a pre-processing step based on extra priors, this paper is mainly considered on the modeling level, proposing the LAD-PINN to deal with PDE reconstruction or coefficient identification problems involving high noise or anomalous data, as well as proposing a two-stage framework improve the accuracy further.

At the algorithm level, two issues remain to be addressed. One is the problem of loss weight selection when it is transformed into an unconstrained problem. However, no systematic methods exist to fix weights for varying noise scales. Fortunately, in the examples of this paper, weights of the losses have extensive acceptable ranges. 
The choice of weights may be avoided by using penalty methods or augmented Lagrangian multiplier methods, although both methods introduce more algorithmic hyperparameters while eliminating model hyperparameters.
On the other hand, we did not specifically consider the non-smoothness of the LAD loss term in the computational sense. 
The loss term might be handled using reweighting methods like in robust analysis\cite{pitselisReviewRobustEstimators2013}, which transform a non-smooth problem into sequential smooth subproblems. Of course, this approach also introduces more hyper-parameters, which are avoided as much as possible in this paper.

\section*{Acknowledgement}
This work was supported by National Natural Science Foundation of China under Grant No.52005505 and 11725211.

\bibliographystyle{elsarticle-num} 
\bibliography{cas-refs}





\end{document}